\documentclass[letterpaper,english,reprint, aps,superscriptaddress]{revtex4-2}
\usepackage{babel}
\usepackage[T1]{fontenc}
\usepackage[latin9]{inputenc}
\usepackage{mathrsfs}
\usepackage{amsmath}
\usepackage{amssymb}
\usepackage{graphicx}
\usepackage[title]{appendix}
\usepackage{xcolor}

\makeatletter

\pdfpageheight\paperheight
\pdfpagewidth\paperwidth

\makeatother

\begin{document}

%

\title{Self-learning Machines based on Hamiltonian Echo Backpropagation}

\author{Victor Lopez-Pastor}
\affiliation{Max Planck Institute for the Science of Light, Erlangen, Germany}

\author{Florian Marquardt}
\affiliation{Max Planck Institute for the Science of Light, Erlangen, Germany}
\affiliation{Physics Department, Friedrich-Alexander-Universit\"at Erlangen-N\"urnberg, Erlangen, Germany}

\date{\today}

\begin{abstract}
A physical self-learning machine can be defined as a nonlinear dynamical system that can be trained on data (similar to artificial neural networks), but where
the update of the internal degrees of freedom that serve as learnable
parameters happens autonomously. In this way, neither external processing and feedback nor knowledge of (and
control of) these internal degrees of freedom is required. We introduce
a general scheme for self-learning in any time-reversible Hamiltonian system. It relies on implementing a time-reversal operation and injecting a small error signal on top of the echo dynamics. We show how the physical dynamics itself will then lead to the required gradient update of learnable parameters, independent of the details of the Hamiltonian. We illustrate the training
of such a self-learning machine numerically for the case of coupled nonlinear wave fields and other examples.
\end{abstract}
\maketitle

\section{Introduction}

In the last decade, the field of Machine Learning (ML) has experienced an explosive growth, finding use in an ever-increasing number of applications in our everyday lives, from automatic driving to face recognition. To a large degree, this astonishing progress in ML can be attributed to the developments in Artificial Neural Networks (ANN). Training deep ANN's has only recently become possible in practice \citep{goodfellow2016}, thanks both to the availability of large data sets and to the continued improvement in digital electronic hardware and the advent of fast Graphical Processing Units (GPU) and other specialized hardware. However, while the demand for faster and more efficient information processing will only grow  in order to address the needs of increasingly larger and complex ANN's, the exponential growth in the power of electronic devices that we enjoyed in the last half century appears to be coming to a halt. 

What is more, the von Neumann architecture that is currently employed by electronic devices is known to be highly inefficient for most ML applications. In a von Neumann computer, the memory and processing units are separated, and the necessary transfer of data between them can severely constrain the overall performance. The field of \textit{neuromorphic computing} \citep{schuman_survey_2017,markovic2020physics} aims to improve the efficiency of specialized hardware for ML by imitating the structure of biological neural networks. The hope is to realize devices that are as efficient and massively parallel as the brain, while using much faster physical processes to carry out the information processing. 

In particular, the idea of constructing neuromorphic computing devices based on light has recently attracted a lot of attention, as it promises to unlock all the benefits of optical computing: broad bandwidth, small latency, low power consumption and natural parallelism \citep{shastri_photonics_2021}. Apart from optics, other physical platforms have been considered too, such as spin-based devices or  memristor circuits \citep{schuman_survey_2017}. 

\begin{figure}[h]
\includegraphics[width=8cm]{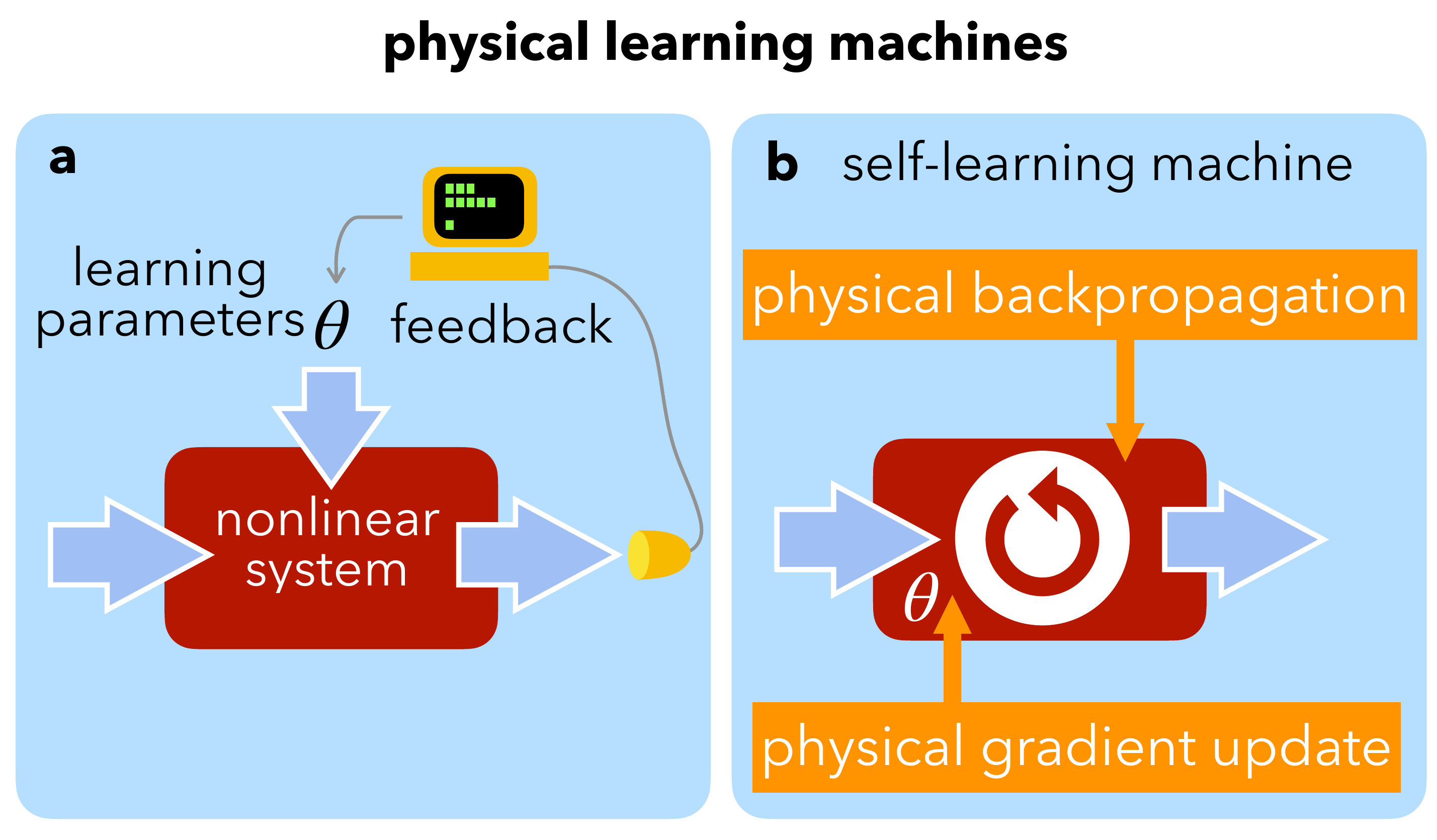}
\caption{\label{Fig1_Schematic} \textbf{Different types of physical learning machines.} (a) Physical learning machine requiring feedback based on the outcome, to tune the internal "learning" parameters $\theta$ inside the physical device. (b) A self-learning machine does not involve feedback. It updates the learning parameters in a fully autonomous way (for example using physical backpropagation and a physical gradient update; see main text). }
\end{figure}

Nonetheless, \textit{training} such machines efficiently is still a challenging problem. The simplest method, shifting each learning parameter individually and providing feedback based on the outcomes, is very inefficient, scaling linearly in the number of parameters. Recently, a method using a combination of a physical forward pass but digitally simulated backpropagation was used successfully, but it requires relatively faithful models of the physical dynamics \citep{mcmahon}.  In another recent line of development, the concept of equilibrium propagation was introduced \citep{scellier2017equilibrium}, leading to a local update rule for adapting weights based on a system's response under externally forced conditions, and it was recently pointed out how to implement such ideas in electronic systems by designing suitable components \citep{martin2021eqspike}.

The best solution would be to realize a \textit{self-learning machine} - i.e. a learning machine that is trained by means of entirely autonomous physical processes (Fig.~\ref{Fig1_Schematic}), without the use of feedback for updating the learning parameters and without any kind of external processing of information (except possibly that needed for feeding the training data). A first step in this direction was already suggested theoretically in a visionary paper by Psaltis et al. \citep{wagner_multilayer_1987} and later implemented to some degree \citep{li_optical_1993,wagner_optical_1993,skinner_neural_1995}.  In Psaltis et al. \citep{wagner_multilayer_1987}, it  was shown that it could be possible to approximately realize the backpropagation algorithm \citep{rumelhart_learning_1986}  in an optical neural network based on volume holograms. However, the nonlinear elements must be engineered so that the transmittance for the backpropagating signal matches the derivative of the transmittance for the forward propagating signal, besides requiring a carefully designed geometric arrangement \citep{psaltis_adaptive_1988}. To this day, such stringent requirements have prevented a fully developed practical implementation of the backpropagation algorithm in an optical learning machine \citep{wetzstein_inference_2020}. 

Other learning machines based on Hebbian learning or Spiking Timing-Dependent Synaptic Plasticity (STDSP) would also fall under our definition of a self-learning machine, and restricted versions have been implemented, e.g. optically \citep{feldmann_all-optical_2019}. Still, both  Hebbian learning and STDSP are motivated mainly by their plausibility in the context of biological neural networks, and it has to be seen in each case how well they optimize some training objective. In soft-matter systems, ideas based on equilibrium propagation have recently been explored \citep{stern2021supervised}.

In this work we consider SL machines based on classical Hamiltonian systems. We present a new training procedure based on time reversal operations, which we name Hamiltonian Echo Backpropagation (HEB). As opposed to existing approaches, we do not need to consider a particular implementation of a SL machine or carefully selected nonlinear elements; instead we present a completely general procedure to train SL machines in a wide class of physical systems.  Indeed, HEB can realize gradient descent and update of the learning parameters via the dynamics in \textit{any} time-reversible Hamiltonian system, and it does so in an efficient way that exploits the parallelism of the learning machine. This new training procedure opens up many exciting possibilities, making it possible to construct self-learning machines in a wide range of physical platforms. One the one hand, HEB makes it feasible to realize SL devices based on already mature technologies, such as integrated photonic circuits. On the other hand, it expands the field of SL machines to new interesting physical platforms, such as clouds of cold atoms or trains of optical pulses in a fiber loop. Given how broad are the sufficient conditions for HEB to work, we believe that it will instigate the discovery of completely new physical learning machines. Moreover, our training procedure is independent of the Hamiltonian that describes the dynamics of the learning machine. Surprisingly, one does not even need to know the dynamics of the physical device in order to train it. 

In the following, after defining self-learning machines, we will introduce the Hamiltonian Echo Backpropagation procedure and present a mathematical proof of its core element. We will then discuss the ingredients of such a machine, illustrate the learning approach in two numerical examples, and finally comment on the prospects for different physical platforms.

\section{Self-learning machines}

\label{Sec-SLM}

\subsubsection*{Definitions: Physical learning machines vs. physical self-learning
machines}

A \emph{physical learning machine} can be defined as a physical device
provided with an internal memory that can process an input to produce
an output, such that the functional dependence between the input and
the output is parameterized by the state of the internal memory. For
example, a physical learning machine can be a photonic circuit made
of beam-splitters, nonlinear elements and variable phase shifters.
In this example, for any input signal that enters the circuit, the
output will depend on the configuration of the phase shifters, which
in this case represent the internal memory. In a physical learning
machine, training is the process of finding the optimal state of the
internal memory to realize some desired input-output relation.

There are several possible ways of training a physical learning machine.
It could be done entirely externally (using numerical simulations)
or by employing feedback, i.e. external processing of the machine's
output to adapt the parameters. However, this step may spoil some of the advantages of using physical dynamics. Going beyond that, the most advanced version would consist in a machine that uses an internal physical process for training. 

We may thus define a \emph{self-learning machine} as a physical learning machine that, when presented a data set, can train itself in a fully autonomous way. During training, a self-learning machine receives a sequence of inputs and some information about the target outputs. It is also legitimate to realize a preset sequence of external operations on the SL machine or to supply energy. What is not permissible is to give any kind of feedback dependent on the internal state of the device. In this sense, a SL machine is a black box: the user provides an input and obtains an output, but requires neither knowledge of nor access to the internal degrees of freedom.


In principle, physical (self-)learning machines could be used for any machine learning task. It is customary to classify machine learning tasks in three broad categories: \textit{supervised learning}, \textit{unsupervised learning} and \textit{reinforcement learning}. In supervised learning, the task is to infer a function that maps an input to an output based on labeled training data, which consists of example pairs of an input object and an output (usually called the \textit{label}). In unsupervised learning,  the goal is to find patterns or model the probability distribution of the training data set, which unlike in the case of supervised learning is not provided with labels. Finally, reinforcement learning comprises all tasks in which an intelligent agent has to take actions in an environment in order to maximize some reward. In this paper we concentrate on supervised learning, although there may also be exciting venues in the application of physical learning machines to unsupervised  or reinforcement learning. Therefore, the self-learning machines that we consider throughout this paper are those that can learn autonomously when they are provided with pairs of input-output data. 

To avoid confusion, we should repeat that in our manuscript "self-learning" refers to an autonomous physical training procedure, as opposed to the computer science terms, "unsupervised", "self-supervised" or "reinforcement" learning, which refer to the type of training task.

\subsubsection*{Different types of physical learning}




The idea of constructing specialized physical hardware for machine learning, i.e. neuromorphic devices, has been
explored in various physics-based platforms \citep{schuman_survey_2017,markovic2020physics}. What we define as physical learning
machines have been studied theoretically and for some cases already demonstrated
experimentally in the context of photonic hardware \citep{li_optical_1993,wagner_optical_1993,shen_deep_2017,guo_backpropagation_2020},
memristor circuits \citep{hu_associative_2015,soudry2015memristor,xia2019memristive}, superconducting circuits \citep{schneider2022supermind}, and spintronic devices \citep{sengupta_spin-transfer_2015,torrejon2017neuromorphic,grollier2020neuromorphic}, among others. However, as pointed out above, learning is still a challenge. In many platforms, the focus has been on demonstrating individual components and ingredients believed to be useful for both inference and learning, or on implementing larger-scale devices, but without in-situ training.

The space of potential learning rules is large. We may subdivide it in several ways. For our purposes, the two most significant questions are whether any given learning approach is based directly on \emph{optimization} and whether the \emph{learning rule is local}. Importantly, there is an overlap between these two categories, and we will be interested in that overlap region, since it promises the ability to obtain learning updates by physical dynamics (local rule) and at the same time have a rigorous mathematical foundation (being based on optimization). Another question is whether the update of the learning parameters requires \emph{external feedback and processing} or can be \emph{implemented by physical dynamics autonomously}.


There is a broad category of neuromorphic devices  that makes use of learning rules inspired by biological neural networks. This corpus of ideas started historically with Hebbian learning and it can be summarized in the motto 'neurons that fire together, wire together'. A more sophisticated version in the context of spiking neurons is the so-called spike-timing-dependent synaptic plasticity (STDP). This has been suggested as a learning mechanism in several neuromorphic hardware platforms, e.g. for spintronic devices (see simulations in \citep{vincent2015spin}), for memristors (see implementation of a building block in \citep{jo2010nanoscale}), and for superconducting circuits \citep{schneider2022supermind}, and even optoelectronic-superconducting hybrid systems \citep{shainline2017superconducting}. Importantly, these learning rules are always local in order to be biologically plausible (i.e. the update in the learning parameters is a function of only the nearest-neighbour neurons). However, the simplest versions of the Hebbian learning rule are in general not guaranteed to minimize some cost function, and they show a poor performance when applied to deep learning architectures. 

It is, therefore, remarkable that there have been recent discussions of biologically inspired learning algorithms that have been shown to approximately
or exactly minimize a cost function that measures the deviation from the desired behaviour. Thus, such approaches would fall in the particularly interesting region of overlap we mentioned above. Sometimes, these are more abstract conceptual ideas (like the free-energy principle of the brain \citep{friston_free-energy_2010}), while other approaches may be more concrete, relating to particular situations like recurrent networks of a certain structure \citep{bellec_solution_2020}. There are interesting models that combine local rules with top-down signals in biologically plausible models, whose learning dynamics approximates backpropagation under certain conditions \citep{sacramento2018dendritic}.
The area of backpropagation-like mechanisms that may be operating in the brain has been reviewed recently in \cite{lillicrap2020backpropagation}.
In general, though, all these approaches, while very interesting, are not straightforward to implement physically in simple neuromorphic devices and their hardware implementation would presumably require careful design and, often, tailored electronic components.

Optimization of a cost function is of course at the core of deep learning in artificial neural networks. There, the required gradients of the cost function with respect to the learning parameters inside the network can be efficiently calculated using the backpropagation algorithm. Unfortunately, this algorithm is nonlocal. The update of any weight has a complicated nonlocal dependence on any other neuron in the network. It is therefore not a priori obvious how to implement backpropagation physically.

In the domain of machine learning, the concept of so-called contrastive learning is at the basis of local update rules that could replace backpropagation. In general, for contrastive learning the learning update is the difference between two contributions obtained in different learning phases. This was first employed to suggest an alternative to backpropagation in so-called 'learning by recirculation' \citep{hinton1987learning} and then more recently generalized and extended, leading eventually to the idea of equilibrium propagation \citep{scellier2017equilibrium}. In equilibrium propagation, one considers a system which relaxes into some steady state, given external forces that represent inputs and outputs for a given training sample. Two learning phases are distinguished, where the outputs are free or are nudged towards their desired values. The difference between the network activations in these two phases defines the weight update, in a local way. This procedure overcomes many  problems of earlier similar ideas, and it implements backpropagation for the optimization of a cost function. While it was first suggested as a principled, local alternative to backpropagation, without immediate ideas on hardware realizations, there has been recent work on possible implementations of this (and related) ideas, both in electronic circuits \cite{martin2021eqspike} and in soft-matter systems \cite{stern2021supervised}. In contrast to the method we will describe, the basic assumption here is always that the system relaxes to an equilibrium steady state (or begins to relax towards that state), for each presented training example.

\begin{figure}[h]
\includegraphics[width=8cm]{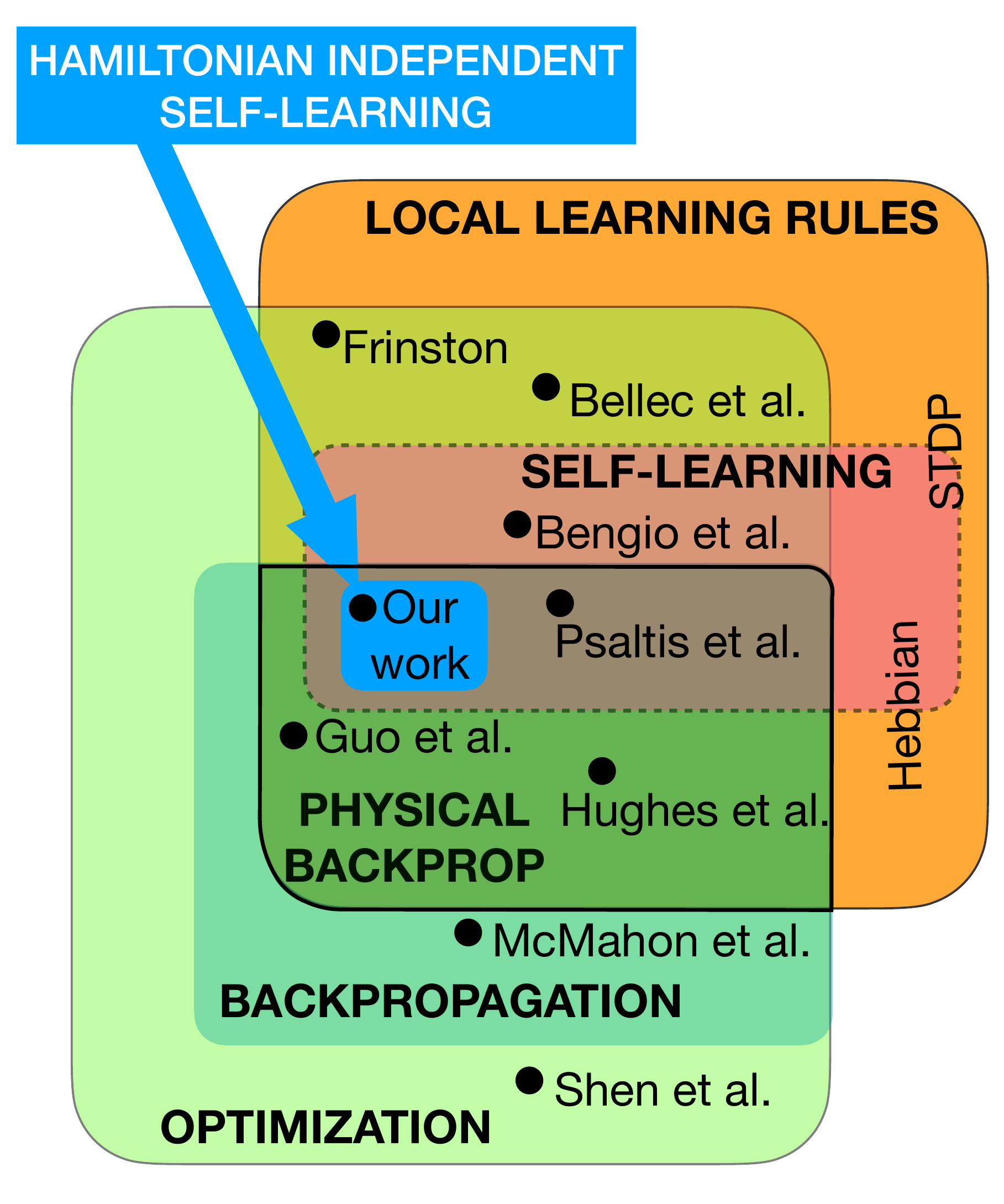}
\caption{\label{Fig2_Overview} \textbf{Overview: The landscape of physical learning machines.} Our work falls into the broad class of optimization approaches. It belongs to the small class of works that involve both physical backpropagation and self-learning, but in contrast to other approaches our scheme is universal, i.e. independent of the specific Hamiltonian. (some related works, see main text: Shen et al \cite{shen_deep_2017}, Guo et al \cite{guo_backpropagation_2020}, Skinner et al \cite{skinner_neural_1995}, Psaltis et al \cite{wagner_multilayer_1987,wagner_optical_1993}). }
\end{figure}


\subsubsection*{The problem of training}

In some of the earlier attempts (like \citep{skinner_neural_1995}), training
was performed \emph{in silico, }by means of numerical simulations. The
problem with this approach is that the size of the learning machine
is limited by the computational power required to simulate the physical
hardware. Moreover, training is performed for an idealized model of
the actual physical hardware. As a result, fabrication imperfections
cannot be taken into account. This has been addressed recently by introducing a hybrid approach, combining physical forward propagation on actual hardware with backpropagation using simulation \citep{mcmahon}.

A general approach to remedy this shortcoming is to employ feedback.

The simplest feedback-based training technique uses a finite-difference
approximation to the gradient of the cost function. This can be realized
physically in a very simple manner: (1) estimate the cost function
$C$ for the current state of the learning parameters, (2) change
the $i$-th parameter by a small amount and estimate $C$ again, (3)
add the difference of these results to the $i$-th parameter. 
However, this idea is far from optimal, since the number of evaluations
needed scales with the number of parameters.

\subsubsection*{Towards physical self-learning machines}

As described above, ideally we would want to both obtain the required training update of the learning parameters and eventually also implement that update using only autonomous physical dynamics.

In this subsection we are going to review those previous developments along this line that we feel are most closely related to our idea. They happen to be in the optical domain -- not surprisingly, given the coherent, time-reversible character of wave propagation.

We know that the backpropagation algorithm that forms the cornerstone
of artificial neural network training can extract the required parameter update (the gradient of the cost function)
using only two evaluations of the network, independent of the number of parameters.
A crucial improvement of a physical learning machine therefore consists in realizing the backpropagation algorithm by physical means, even in cases where the eventual update is implemented by other means. This idea
was first introduced in Wagner et al. \citep{wagner_multilayer_1987},
in the context of optical neural networks. In a typical optical neural
network, an optical field propagates inside some nonlinear medium
with elements that provide controllable phase shifts. The controllable
phase shifts play the role of the learning weights. The optical field
would be injected at the input and propagate towards the output. The
idea advocated in \citep{wagner_multilayer_1987} then is to create
a physical \emph{error signal} that propagates in the opposite direction,
from output to input. The error signal is prepared according to the
difference of the output of the device and the desired target output.
Without going into any further detail (we refer to the original paper),
in their approach the backward-propagating weak error signal interferes
with the strong forward-propagating optical field. In this way, for
the specific choice of optical nonlinearities in their setup it was possible
to ensure that the error signal is (approximately) equal at any point of the device to the required gradient of the cost function. 

Once the error signal
is prepared, one can measure it and use the result to update the parameters
via feedback. All learning parameters could be updated at once, in
a fully parallel fashion. This idea was originally proposed for optical
neural networks using Kerr nonlinearities \citep{wagner_multilayer_1987},
but it was recently extended to setups that employ saturable
absorption for the activation function \citep{guo_backpropagation_2020}.
In a similar spirit, in Hermans et al \citep{hermans_trainable_2015},
the idea of physical backpropagation is used to train a physical linear
system with controllable nonlinear feedback. An optical implementation based on this concept was demonstrated in a follow-up paper \citep{photonicdelay2015}. A related idea by Hughes,
Fan et al. \citep{hughes_training_2018} uses an auxiliary circuit
to backpropagate the error signal. A recent paper proposes a training method that combines in-situ measurements with digital backpropagation \citep{mcmahon}. Other implementations only train the last layer, in a similar fashion to reservoir computing.

Nonetheless, there is a potential drawback in using feedback to update
the learning parameters. The error signal has to be measured in order
to update the learning weights. If that is done by means of electronic
sensors, that could potentially introduce a bottleneck. The potential
advantage of using ultrafast physical dynamics for evaluating the
network could be lost because of a relatively slow electronic response.
In addition, some of these approaches require further computation
steps. These problems can be addressed if the error signal could directly influence the learning parameters via physical dynamics, without external feedback.

The possibility of using a physical mechanism to update the learning
parameters autonomously was proposed for the first time also in the above-mentioned work by Wagner
and Psaltis \citep{wagner_multilayer_1987}. In their paper, they
consider a setup in which the learning parameters are recorded in
the form of a hologram in a photorefractive material. In this way,
the interference between the error signal and the forward-propagating
beam could in principle provide the means to update the holographic
recording in the right way. A simple version was partially demonstrated
in a linear optical device in a follow-up paper by Li, Psaltis et
al \citep{li_optical_1993}. However, that experimental setup did
actually still involve some form of feedback.

The ideas first advanced in \citep{wagner_multilayer_1987} could potentially
lead to ultrafast training and a huge density of interconnections
between neurons. By combining physical backpropagation with an autonomous
process to update the parameters according to the error signal, it
would be possible to realize self-learning machines. Nevertheless,
all presently existing proposals for physical backpropagation are
suitable only for some setups with very specific nonlinearities. In
some cases, such as \citep{wagner_multilayer_1987} or \citep{guo_backpropagation_2020}, it is required that
the forward propagating beams and the error signal experience a different
dynamics at the nonlinear elements, that nonetheless has to be matched
in a precise way.  What is more,
recording the learning parameters in a holographic pattern requires
a very careful engineering of the geometry of the device \citep{psaltis_adaptive_1988}.
In other cases, such as \citep{skinner_neural_1995}, physical backpropagation
is only realized approximately in a particular limit.

In this work, we show how to realize self-learning (i.e. the combination
of physical backpropagation and autonomous parameter update) in a
much broader set of physical systems. Since our idea does not rely
on a specific choice of the system, it is not restricted to optical
setups. Hence, it could be applied to many other possible platforms,
ranging from cold atom clouds to solid state spin networks. In fact,
the training procedure that we propose is a sequence of operations
that do not even depend on the particular dynamics of the device,
provided that some weak assumptions are met. Essentially, we only
require that the physical learning machine is based on a time-reversible nonlinear
Hamiltonian system.

\section{Overview of Hamiltonian echo backpropagation}




The main contribution of this paper is to introduce a new technique for self-learning in Hamiltonian systems, which we term Hamiltonian Echo Backpropagation (HEB, for short). Before moving to the more technical sections, we provide here a high-level overview of HEB.

\subsection{Requirements}

Let us be more precise about the requirements for the implementation of a self-learning machine based on HEB. 

{\bf Hamiltonian systems} -- First
of all, we only consider Hamiltonian systems,
where the dynamical degrees of freedom consist in the learning parameters
and the variables needed to process the information. From the point of view of the physical realization, requiring Hamiltonian dynamics means that
the timescale of dissipation in the self-learning device must be much
larger than the time interval between the injection of the input and
the production of the output (for a given training sample). Since we aim to use fast dynamical processes,
this seems to be a reasonable requirement. In fact, many of the previous
proposals for physical learning machines can be modeled as Hamiltonian
systems, e.g. the optical neural networks using Kerr nonlinearity.

%

{\bf Time-reversal
symmetry} -- Second, we require that the self-learning device obeys time-reversal
symmetry. 
It is worth noting that time-reversal symmetry is not a very restrictive property: all mechanical and optical systems with arbitrary nonlinear dynamics are time-reversal invariant in the absence of magnetic fields and in the absence of spontaneous time-reversal symmetry breaking (i.e. magnetization).

For the purposes of our paper, we will define a Hamiltonian self-learning machine to be a self-learning machine that
is Hamiltonian and time-reversible.

{\bf Time-reversal operation} -- Third, we need the ability to physically implement a time-reversal operation. In the case of wave fields, that would be a phase-conjugation mirror, an operation which has been demonstrated in nonlinear optics. This can be implemented outside the core of the device.

{\bf Decay step} -- Fourth, we need the ability to implement another operation, the so-called decay step, to be explained below. Essentially, this is when the force imparted on the learning variables gets converted into a shift of those variables. Again, this can be implemented outside the core of the device.

These are the only general requirements needed for the implementation of self-learning machines based on the Hamiltonian Echo Backpropagation technique. Any other more detailed suggestions regarding physical implementations made in the remainder of the manuscript only serve to highlight opportunities for optimizing the performance -- similar to discussions in deep learning where one should distinguish the essential concepts, i.e. gradient descent and backpropagation, from suggestions for improved network architectures.


\subsection{Basic steps of Hamiltonian echo backpropagation}


We explain here all the steps needed to implement HEB in what is probably the most practical scenario, which is when wave fields are transmitted through a nonlinear setup. In this setup, wave fields represent both learning variables, to be updated during training, and evaluation variables, participating in the information processing required to map input to output. Both types of wave fields enter a physical system (the \textit{nonlinear core}), in which they interact to produce an output wave field. The only requirement that we impose on the nonlinear core is that the evolution of the fields inside it is described by a time-reversible Hamiltonian. 

Apart from the nonlinear core, the self-learning device requires a few well-defined external operations that are applied on the wave fields that have been transmitted through the device. However, we need no control on the nonlinear core, which can be thought of as a black box. Since all the external operations depend only on the  output fields and not on the internal degrees of freedom or the details of the dynamics inside the nonlinear core, we will call this a Hamiltonian-independent self-learning machine.

Suppose that we consider the usual supervised learning scenario, with a dataset consisting of pairs of inputs and  target outputs, and we want to train the device to approximate the input-output relation given by these samples. We proceed to enumerate the steps in a single training iteration of HEB.

1) The first step is to draw a random sample from the input dataset and prepare the evaluation field accordingly. 

2) \textbf{Forward pass:} send both the evaluation field and the learning field into the nonlinear core and let them interact there. This is where the essence of the nonlinear information processing (and, later, learning process) will take place. 

3) Both fields have again emerged from the system, propagating outwards as wave fields. We now inject a small \textbf{error signal} on top of the evaluation field. This signal is proportional to the desired output of this training sample.
Note that this prescription covers the case of the most well-known cost function, the mean square loss, and it corresponds mathematically to injecting a signal proportional to the gradient of the cost function with respect to the output field. Other cost functions can be implemented as well, as we will show.


4) \textbf{Time-reversal Operation:} This is performed by phase-conjugating both wave fields, which at this point are outside the nonlinear core. More generally, it means reversing the momenta of the physical degrees of freedom. 

5) \textbf{Backward pass:} As a result of the time-reversal operation, the fields will propagate back into the nonlinear core, where they will interact again (we will call this the \textit{backward pass}).

(6) Now the evaluation field leaves the device near the input port (it will be substituted by the next input). However, the learning field is retained, and a \textbf{second time-reversal operation} is now enacted on this learning field, again outside the device. This will eventually re-inject the learning field into the device, together with the next input sample.

7) \textbf{Decay step:} In a Hamiltonian setting, the degrees of freedom of any field can be decomposed into the field amplitude and its canonically conjugate momentum; for the optical case, these correspond to the two field quadratures. As we shall prove in our work, the canonical momentum of the learning field after the steps described above is proportional to the gradient of the cost function, i.e. exactly what is needed in a gradient-descent learning algorithm! The purpose of this final step is to convert the canonical momentum (i.e. the field quadrature whose sign we have flipped during the time-reversal operation) into a change of the field amplitude and quench the momentum to zero in the process. We call this the decay step, since it is the only step in which dissipation is introduced. As this step takes place outside the nonlinear physical system (just like the time-reversal operation itself), it can be realized in an engineered way without destroying the Hamiltonian-independence advertised above. Furthermore, since it is realized outside the nonlinear core one can control the learning rate in a simple manner.

We will discuss all of these steps in the following and provide proofs for the claims made here. Nevertheless, we can already offer an insight of why the full procedure works. Consider that we wanted to evaluate the gradient by the finite-difference method. In this case, we would have to slightly shift one of the learning fields and query the resulting response. Such a shift would result in a weak perturbation of the fields during the forward pass that would ultimately result in a small change of the value of the cost function with respect to the case of unperturbed fields. Now we would have to repeat this process for every learning degree of freedom to obtain the full learning gradient, which is one of the reasons why, as we explained above, the finite-difference method is hugely inefficient. But since we consider systems with time reversal symmetry, we can instead inject an error signal in the output, followed by a time-reversal operation. In this case, the wave fields evolve to a time-reversed replica of the input that is perturbed by a weak error signal. Because of the reciprocity principle \citep{stakgold_greens_2011}, the linear response of the (nonlinear!) system in the forward pass is in fact related to the linear response of the system in the backward pass. Therefore, we can obtain the required gradient by injecting a signal from the output, instead of perturbing the input. This is the essence of the idea underlying our approach.


Our method shares with the approach of equilibrium propagation \citep{scellier2017equilibrium} the general idea of exploiting a difference between two contributions: in our case, this would be the difference of the force imposed on the learning field between the forward pass and the backward pass. The difference is automatically produced by the physical dynamics since we employ a time-reversal operation. In contrast to equilibrium propagation, we do not have to engineer further the dynamics inside the device to produce an update according to such a difference. Another distinction  is that we work with a system out of equilibrium where we do not have to wait for it to relax to any steady state (nor would we want that, since our method relies on dissipation being absent or at least very small). 

Our method shares with the approach of Psaltis and Wagner \citep{wagner_multilayer_1987} the idea of employing a time-reversal operation. However, in our case, we apply time-reversal to the full nonlinear wave field, as opposed to only an injected weak error signal. This difference ultimately means that they had to engineer the forward and backward transmissions inside the device components in a particular way, while our approach is completely general. 

We observe at this point that none of the external procedures that are employed during an HEB iteration depends on the Hamiltonian of the nonlinear core. Therefore, we introduce the following definition. A Hamiltonian-independent self-learning machine is a Hamiltonian self-learning machine in which the training procedure is independent of the Hamiltonian. By this statement, we mean that the training procedure can minimize the cost function for
any Hamiltonian that is time-reversible. This is another important difference to equilibrium propagation as well as the approach of Wagner and Psaltis.

To the best of our knowledge, our proposal constitutes the first example
of an approach to construct Hamiltonian-independent self-learning
machines. Since the training procedure does not depend on the particular
form of the Hamiltonian, the self-learning machine can be regarded
as a black box. As opposed to other physical learning machines in
the literature, its implementation in the laboratory does not even
need a detailed knowledge of the internal dynamics of the device in
order to make it work. There are only few mechanisms that could
disrupt an experimental implementation: a mechanism that in some way breaks time-reversal
symmetry of the Hamiltonian (including an unwanted dissipation
or noise channel) , or significant imperfections in the implementation of the time-reversal operation or the decay step.

\section{Hamiltonian Echo Backpropagation}

\label{Sec-HEB}

\subsection{Dynamical variables and Hamiltonian}

Our setup for a SL machine consists of two basic ingredients:
dynamical variables that are used to process the information (the
\emph{evaluation} variables), and dynamical variables that contain
the learning parameters. 

The evaluation variables will be combined into a single dynamical vector, $\Psi(t)$. In a similar way, the collection of all the learning parameters is denoted as a vector $\Theta(t)$. Such vectors can represent any collection of physical degrees of freedom in a Hamiltonian system, from interacting spins to optical fields. For the present discussion, we imagine discrete models with a countable number of degrees of freedom. However, the approach also covers the case of continuous fields (with straightforward slight changes in the notation).

The input to the device is given by the initial configuration of the
evaluation variables, $\Psi(-T)$, whereas
the output is given by its state at a later time, $\Psi(0)$. We choose the initial time to be $-T$ for reasons that will become clear later.
The learning parameters, $\Theta(t)$, will interact
with the evaluation variables and overall change slowly during training.

For example, consider a self-learning machine where $\Psi$ and  $\Theta$ are nonlinear wave fields. These could be realized as a lattice of optical
cavities or spins. We can
imagine the operation of the SL machine as a nonlinear scattering experiment, where the input $\Psi(-T)$ is a wave packet.  As the wave packet propagates inside the device, it evolves under the effect of the nonlinear self-interaction and the interaction with the learning parameters. Finally, an output wave packet is obtained at $t=0$. The output state of $\Psi$
is a nonlinear function of the input, and it has a parametric
dependence on the initial configuration of the learning parameters, $\Theta(-T)$, much like an artificial neural network (ANN).




We can capture the dynamics of the SL machine during both the forward and backward pass in a single Hamiltonian:

\begin{equation}
H_{SL} = H_{\psi} + H_{\rm{int}}+H_{\theta}.\label{eq:Hamiltonian_SL}
\end{equation}
Let us inspect one by one what each term means. Some of the technical details will be more fully explained later.

$H_{\psi}$ is the Hamiltonian that describes the evolution
of $\Psi$ alone, in the absence of interaction with $\Theta$. In general, it will have self-interaction (non-quadratic) terms. The second term, $H_{\rm{int}}$, contains the interaction
of $\Theta$ and $\Psi$. In general, $H_{\psi}+H_{\rm{int}}$
must be a time-reversible Hamiltonian, but there is no further
constraint. In accordance with previous proposals for physical learning machines, we may assume
that, for example, $\Psi$ is an optical field inside a nonlinear
medium. The nonlinear response of such a medium may correspond, among
many other possible choices, to a Kerr-like nonlinearity.

The third term, $H_{\theta}$, describes the dynamics of
$\Theta$ in the absence of interaction with $\Psi$. For the sake of generality, we consider that $\Theta$ may have an arbitrary self-interaction. Therefore, $H_{SL}$ is completely arbitrary.

\subsection{Gradient descent as a goal}


Suppose that $\Theta$ is initialized in some random configuration
$\Theta(-T)$. Now, we prepare $\Psi$ in the input state, and we
let it interact with $\Theta$. After a time interval $T$, we will
obtain the output state.  The error in the output is quantified by means of a cost function. In general, a \textit{cost function} can be any function 

\begin{equation}
C(\Psi(0),\Psi_{target}) = C(\Psi(-T),\Psi_{target},\Theta(-T)) \label{CostFunctionEq}
\end{equation}
of the output $\Psi(0)$ and its corresponding target configuration that is minimized when both are equal. The function displayed here is the sample-specific cost function. The training procedure will try to minimize its sample-averaged version (averaging over all training input samples $\Psi(-T)$ that are provided alongside their respective target $\Psi_{target}$). After averaging, the cost function still depends on the learning variables $\Theta(-T)$, since these determine the mapping from input to output. 

For reasons that will become apparent later, we take our cost function to have units of energy (and in fact it will represent a physical Hamiltonian later in this section). In order to improve the performance of the SL machine, we have to minimize the
cost function.  In principle, the problem of optimizing the cost function is a very hard nonlinear optimization problem, but in the context of ML it is usually sufficient to find a local minimum. In the context of ANN's, this is normally done by the
stochastic gradient descent method. In this method, the gradient
of the cost function is computed (for one or several randomly chosen training samples) and the learning parameters are subsequently updated, according to the rule

\begin{equation}
    \theta(T_f)=\theta(-T)-\eta\frac{\partial C}{\partial \theta(-T)} .
\end{equation}

\subsection{Introducing Hamiltonian Echo Backpropagation}

Our goal is to realize the stochastic gradient update rule
in a fully autonomous way, via physical dynamics. The sample-specific cost function depends on the output
and the target state. For this reason, if one hopes to implement gradient
descent without external intervention, some information about the
output must be sent \emph{backwards} from the output to the input.
This immediately suggests the use of a time reversal operation, which was already recognized in early proposals \cite{wagner_multilayer_1987}, although it was employed in a manner different than what we will describe now. 

To understand the qualitative idea of our approach, we start from a general observation that is well known from discussions of backpropagation in ANNs. The change of the cost function induced by a small change in a learning parameter $\theta$ is given by

\begin{equation}
\delta C = 2\delta \theta Re\left\{\frac{\partial \Psi(0)}{\partial \theta} \frac{\partial C}{\partial\Psi(0)}\right\}
\end{equation}

In other words, we need to know both the perturbation in the output
configuration $\delta\Psi(0) = \frac{\partial \Psi(0)}{\partial \theta}$ produced by a small parameter change, as well as the "error signal" $\frac{\partial C}{\partial\Psi(0)}$ 
at the output. 

\begin{figure}[h]
\includegraphics[width=8cm]{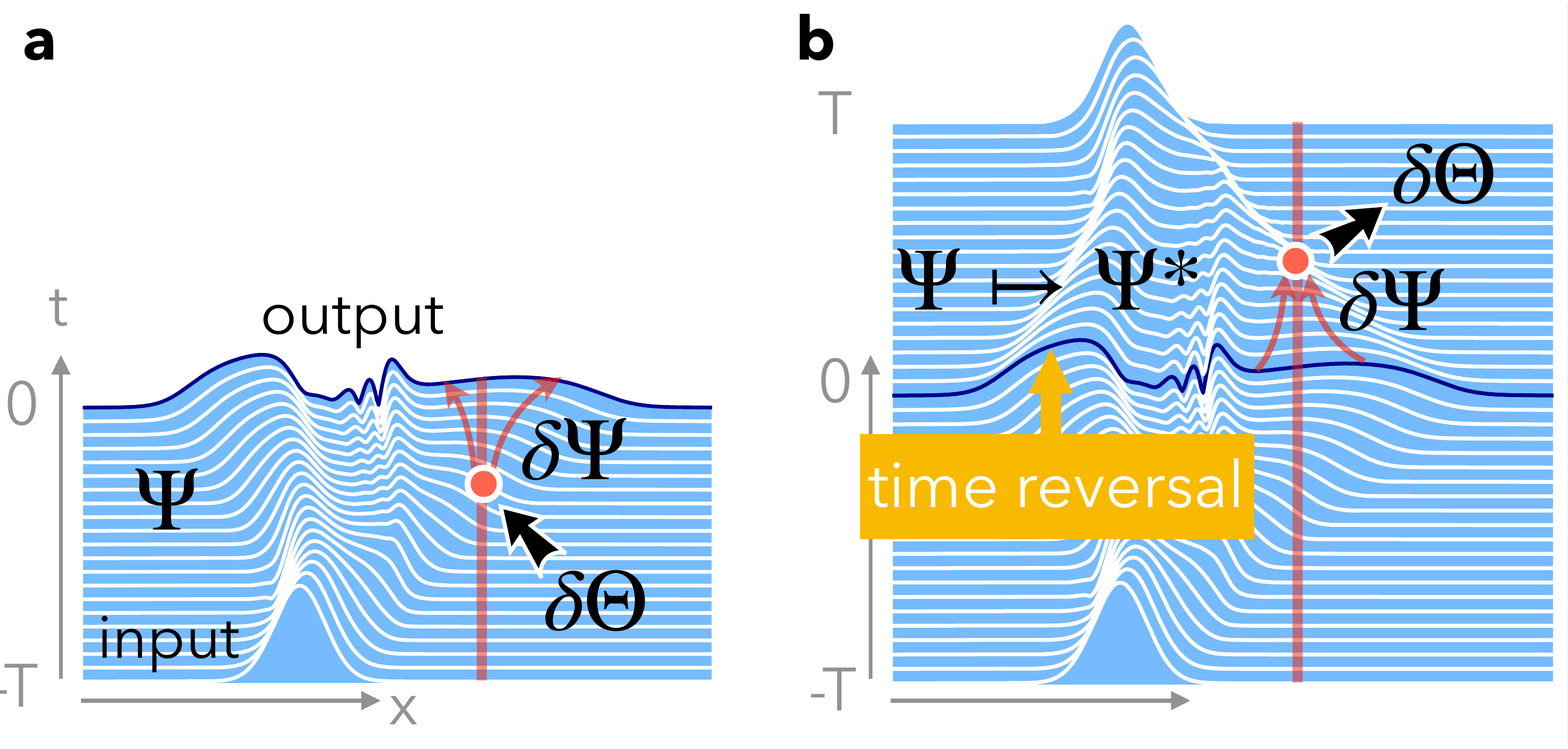}
\caption{\label{Fig3_Backprop} \textbf{Physical backpropagation and physical gradient update.} (a) Forward pass (nonlinear evolution) of the "evaluation" wave field $\Psi$ in the presence of a background potential landscape provided by a "learning field" $\Theta$. Any small perturbation $\delta \Theta$ will affect the output, as indicated here, which in turn changes the value of the cost function.  (b) Instead of testing the cost function response for many potential updates of the learning field $\Theta$, Hamiltonian Echo Backpropagation automatically produces the required update in a single backward pass, via physical dynamics. After applying a time-reversal operation ($\Psi \mapsto \Psi^*$), the subsequent nonlinear wave field evolution is a time-reversed version of the forward pass. A small variation $\delta \Psi$ injected on top can now be used to backpropagate the error signal (variation of the cost function at the output); see main text. This then leads to the required update of the learning field $\theta$, via the physical interaction between $\Psi$ and $\theta$, realizing a fully autonomous learning machine. }
\end{figure}

Since we consider weak perturbations, $\delta\Psi(0)$ can be understood in terms
of the linear response of the system. In other words, $\delta\Psi(0)$ is in fact
proportional to the Green's function associated with the linearized
equations of motion. At this point we can introduce the most important 
idea of our approach: if we physically time-reverse the whole field $\Psi$, we can instead interpret
the error signal, $\frac{\partial C}{\partial\Psi(0)}$  as the source of a perturbation that is "riding on top of" the time-reversed nonlinear wave field (see Fig.~\ref{Fig3_Backprop}). This is at the heart of our new procedure to realize
self-learning, "Hamiltonian Echo Backpropagation" (HEB).

As we will see later, the idea is related to the fact that in a linear system of differential
equations there is a symmetry between the source of a field at $t$
and its effect at $t'$ (the so-called self-reciprocity
principle) \citep{stakgold_greens_2011}.

Let us go through the individual steps of the HEB procedure. First, $\Psi$ is prepared in the input state, at time $t=-T$. Note that we
choose to start the process at time $t=-T$ to make explicit the time-reversal
symmetry. Thus, the output is produced at $t=0$, and the echo of the
input will form at $t=T$. In order to simplify the calculations, let
us also assume first that the canonical momentum of $\theta$ is
initially zero. After the interaction between
$\Psi$ and $\Theta$, we obtain an output field $\Psi(0)$. 

In the next instant, we need a way to inject the error signal required for the subsequent backward pass.
This we achieve by imposing on $\Psi$ an interaction Hamiltonian proportional to the
(sample-specific) cost function, given the desired target field configuration. We assume that the duration of this interaction, $\epsilon$, is
small enough. In practice, this implies that the field is slightly
changed according to $\Psi(0)\rightarrow\Psi(0)-i\epsilon\partial_{\Psi^{*}(0)}C$

Before going further in the discussion of HEB, let us introduce a useful
notation we will employ in the rest of the paper. In order to make our equations
more compact, we condense each pair of canonical coordinates in a
single complex variable. We will reserve the capital greek letters
for the complex variables, as follows: $\Psi=\psi+i\pi_{\psi}$, $\Theta=\theta+i\pi_{\theta}$,
where the canonical coordinates and momenta have been rescaled to
obtain the same dimension (nominally, square root of an action). This will make the notation for the time-reversal
especially convenient, as it corresponds to a complex conjugation of
the dynamical variables. Furthermore, one can easily check that Hamilton's equations
can now be written in a particularly convenient manner: 
\[
\dot{\Psi}=-i\partial_{\Psi^{*}}H,
\]
where one has to formally treat $\Psi,\Psi^{*}$ as independent variables
(or more technically, $\partial_{\Psi^{*}}$ represents a Wirtinger partial derivative).

As a concrete example, let us assume that our cost function is simply the overlap between the target state and the output state. Then, the prescription above tells us we just have to inject a weak perturbation proportional to the target state: $\delta\Psi(0)\propto i\epsilon\Psi_{target}$. 

After this step, we time-reverse , which in our complex notation means to phase conjugate the fields, e.g. $\Psi\rightarrow\Psi^{*}$. In the limit $\epsilon\rightarrow0$, $\Psi$ will evolve backwards precisely to the input configuration, because the Hamiltonian is time-reversal-invariant.
However, when $\epsilon$ is small but finite, we obtain a small variation
in the fields, both in $\Psi$ and $\Theta$.  One can understand such
variation as a perturbation traveling on top of the time-reversed
field, from output to input. Once the back-propagation is complete, one obtains a slightly perturbed echo of both $\Psi$ and $\Theta$, at $t=T$. Since we have perturbed the field at $t=0$, the echo is not perfect: there is a small variation given by $\Theta(T)=\Theta^{*}(-T)+\delta\Theta(T)$. In order to have the correct sign for the momentum of the learning field required for our approach, we time reverse again, $\Theta(T)\rightarrow\Theta^*(T)$. Using the self-reciprocity principle, one can then show that the final configuration of $\Theta$ in the echo step is simply given by 
\begin{equation}
\Theta(T) = \theta(-T)-i\epsilon\frac{\partial C}{\partial \Theta^*(-T)}. \label{update_echo}
\end{equation}
We will show below how to prove that result, which encapsulates the central
idea of our proposal,  mathematically (Sec.\,\ref{Sec-Echo-Step-Math}).

We note that the procedure introduced here differs in an important way from previous approaches to self-learning dynamics \cite{wagner_multilayer_1987}, which never considered a full time-reversal of the nonlinear wave field and instead treated a weak perturbation traveling against a strong forward-propagating field. It is a consequence of this choice that these approaches work only for some specific carefully chosen wave dynamics, and not in the general, Hamiltonian-independent way that we outline here.

The final step of HEB will be the update
of the learning field, to be discussed now.

\subsection{The decay step (physical learning field update)}

\label{subsec_learning_field_update}

At this step, the dynamics of the learning field $\Theta$ almost looks like gradient descent, but not completely:
we have been able to impart a momentum kick to $\Theta$ that is proportional to the gradient of the cost function, but what we really want is to update the position $\theta$. We now need to convert the update in the canonical momentum into a shift in the position, and at the same time we need to let the momentum decay to zero. For this purpose, we need to realize some form of dissipation during this part of the training process, which we call the \textit{decay step}. This can be done by coupling $\Theta$ to a reservoir. 

While we need dissipation during the decay step, the dynamics during the forward and backward pass must be time-reversible. The conceptually simplest way to resolve this conflict is to have the ability to switch dissipation on and off. In principle, we can do so by controlling the coupling of $\Theta$ to the reservoir, and many physical implementations (e.g. all approaches related to laser cooling ideas) are possible.  The fact that we need dissipation during the decay step is not surprising, because we have only used Hamiltonian evolution and time reversals so far, while the overall gradient descent procedure is contractive. 

In general, therefore, during the decay step we need to switch on a dissipative interaction 
\begin{equation}
H_{\rm{SL,decay}} = H_{\rm{SL}}  + H_{\theta,\rm{decay}} 
\end{equation}

where $H_{\theta,\rm decay}$ is a term that  couples $\theta$ to the dissipative bath. Additionally, $H_{\theta,\rm decay}$ in general needs to contain other simple non-dissipative terms that control the evolution during the dissipative phase.

We will now explain this step by means of a simplified example that contains all the key ingredients needed. Later on, in section \ref{sec:implementation-decay-step}, we will provide a more detailed prescription how the desired effect outlined here can be obtained in the concrete case of wave fields. 

In our example, we assume that the dynamics of $\Theta$ during the decay step corresponds
to a free-particle Hamiltonian with damping: $$\dot{\theta}=\Omega \pi_{\theta}$$
$$\dot{\pi}_\theta=-\Gamma(t)\pi_\theta$$(Note that in our convention $\pi_\theta^2$ has units of action, which implies that $\Omega$ has units of frequency). 
Furthermore, we assume that the effect of dissipation may be modelled by a damping force of the form $-\Gamma(t)\pi_{\theta}$,
where $\Gamma(t)$ can be controlled in a time-dependent fashion.  The stable manifold would be given by the particle at rest at any location, with $\pi_\theta=0$. 

During this final step of the HEB procedure, we switch on the dissipation, ${\dot \pi_{\theta}} = - \Gamma \pi_{\theta}$, and apart from that we let $\Theta$ evolve freely.  If we wait long enough, the final configuration
of $\Theta$ is given  by $\theta(-T)+\epsilon\frac{\partial C}{\partial\pi_{\theta}(-T)} -\epsilon\frac{ \Omega}{\Gamma}\frac{\partial C}{\partial\theta(-T)}$ (to first order in $\epsilon$), while the canonical momentum decays to zero. The update step finishes in the end of the decay step, at time $T_f\gg\Gamma^{-1}$. In the limit of large decay times, i.e. $\frac{\Omega}{\Gamma}\gg \frac{\partial C}{\partial\pi_{\theta}(-T)} \left(\frac{\partial C}{\partial\theta(-T)}\right)^{-1} $,  we recover the usual gradient descent update rule
\begin{equation}
\theta(T_f)=\theta(-T)-\eta\frac{\partial C}{\partial {\theta}(-T)}, \label{eq_update_free}
\end{equation}
with a learning rate given by $\eta = \epsilon\frac{ \Omega}{\Gamma}$. Repeating this procedure in a sequential way, for many injected training samples, will realize stochastic gradient descent (SGD). Apparently, the learning rate is proportional to $\epsilon$, which represents the duration during which the "cost function Hamiltonian" is active.

We must remark that in the general case, when the dynamics of $\Theta$ is not necessarily equivalent to a free-particle with damping, the relation of $\eta$ to $\epsilon$ may be different. However, we can always expect these two quantities to be proportional, due to the assumption of injecting only a weak perturbation.

\begin{figure}[h]
\includegraphics[width=8cm]{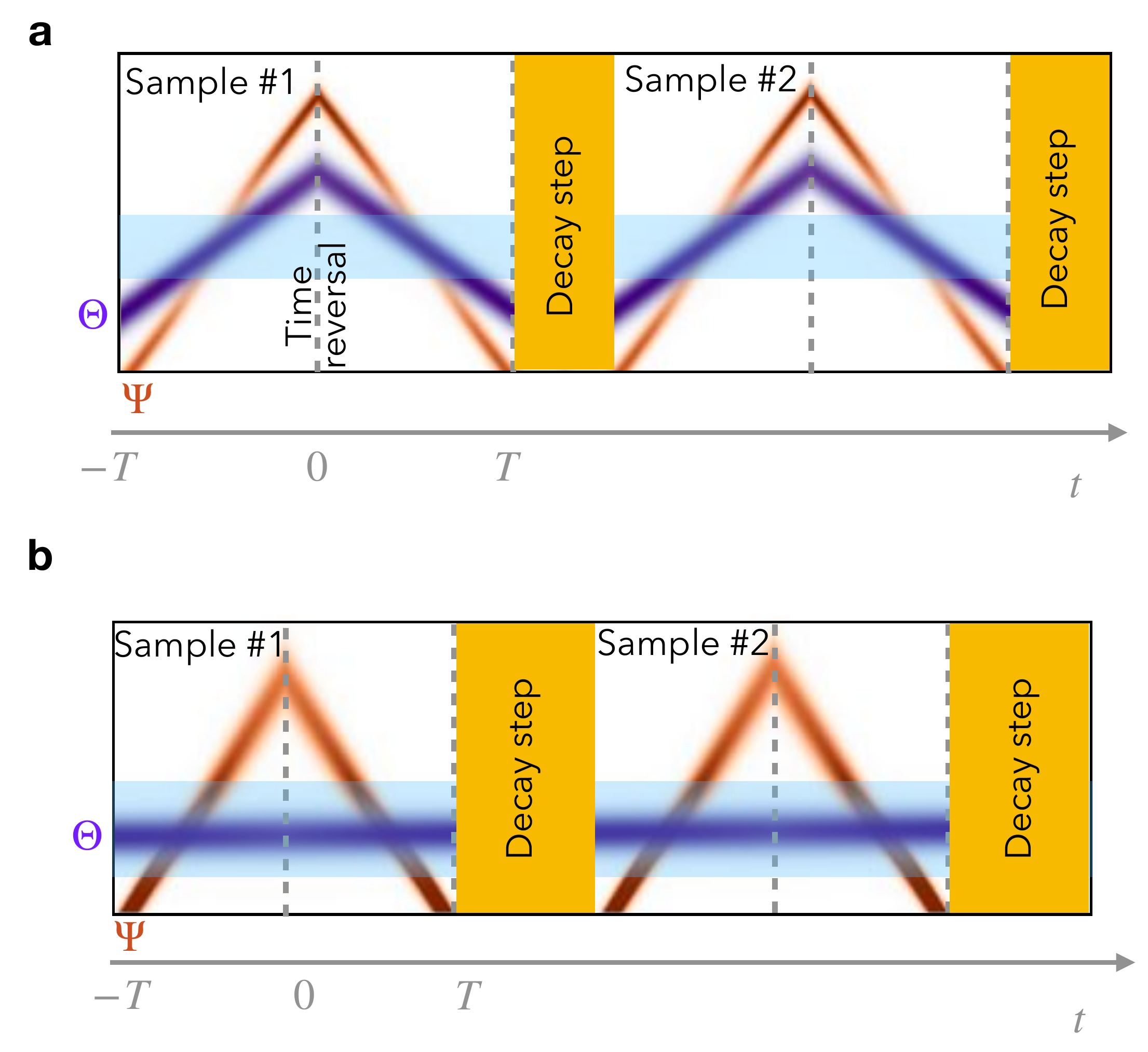}
\caption{\label{Fig_Variants} \textbf{Different possible architectures of the self-learning machine.} We illustrate the different possible configurations of the learning and evaluation dynamical variables, for the case of a setup consisting of wave fields interacting in 1D. We consider that the SL device is in the highlighted region, which is where the nonlinear interaction given by $H_{\rm{SL}}$ takes place. If both the input and output of both fields are injected from outside, all the time reversals and decay steps can be performed externally, and there is freedom to choose $H_{\theta,\rm{bath}}$. If not, the Hamiltonian of the SL device, $H_{\rm{SL}}$, has to be engineered to have a continuously degenerate ground state in $\Theta$.  (a) Both $\Psi(-T)$ and $\Theta(-T)$ are wave packets injected from outside the SL device. (b)  The input $\Psi(-T)$ is a wave packet injected from outside the device, but $\Theta$ is confined inside. }
\end{figure}

In practice, the way in which the decay step is performed would depend on the architecture of the self-learning machine. We can envisage two main possibilities, which are illustrated in figure \ref{Fig_Variants}. 

Before we discuss these possibilities, we emphasize that it is important to have an overall dynamics where the degeneracy between different values of the learning field is only broken by the training samples, and not by the intrinsic dynamics of the device. After all, the learning parameters $\Theta$ must be able to change continuously between different states during training.

{\bf Learning field transmitted through the nonlinear core} -- The first possibility, illustrated in figure \ref{Fig_Variants}(a),  is that both $\Theta$ and $\Psi$ are wave fields that enter in the self-learning device from outside, evolve according to $H_{\rm{SL}}$, and then  finally produce an output wave field. In this case, the echo of both  $\Theta$ and $\Psi$ is produced outside the device. Therefore, all the operations needed to realize of the decay step can be performed externally, and there is freedom to choose the most suitable form of $H_{\theta,\rm{decay}}$. In this situation, the procedure is completely Hamiltonian-independent, since the operations required to implement $H_{\theta,\rm{decay}}$ are the same for any time-reversible Hamiltonian self-learning device. In this case, even when the training is finished, one must keep time-reversing $\Theta$ and performing decay steps in order to keep it around the optimal configuration.
In the absence of an error signal injected at the output, the time-reversed evolution during the backward pass will exactly undo any evolution during the forward pass, regardless of the Hamiltonian of the nonlinear core. Therefore, the learning field will keep stable once the training is finished. The only conditions to impose are on the Hamiltonian governing the dissipative dynamics outside the device, $H_{\theta,\rm{decay}}$, which must not break the degeneracy between different values of $\theta$ (unlike the Hamiltonian $H_{\rm SL}$, which can be arbitrary in this case).

For this architecture, since the decay step is performed outside the device, in principle there is no constraint on the values of the learning rate or the particular form of dissipation. In this case, the learning rate can be tuned to maximize the speed of training. Just like in the case of training of artificial neural networks, the learning rate cannot be too high: otherwise, a gradient descent step is not guaranteed to decrease the value of the cost function.

{\bf Learning field localized inside the nonlinear core} -- The second possibility, illustrated in figure \ref{Fig_Variants}(b), is that the evolution of  $\Theta$ happens entirely inside the device. For example, this would be the case if the learning dynamical variables are given by optical fields confined in cavities placed inside the self-learning device, or are stored inside some other persistent degrees of freedom. In that case, there is the danger that during the decay step the dynamics of prescribed by $H_{\rm SL}$ would induce some unwanted drift of the learning field. To avoid this, we need to require that the dynamics  induced by $H_{\rm SL}$ results in a continuously degenerate ground state manifold for $\Theta$. Therefore, when the learning dynamical variables are confined inside the device, the form of $H_{\theta}$ is constrained. In appendix G we outline some possibilities to engineer such a Hamiltonian with a few simple nonlinear interactions.
Nevertheless, we can still have arbitrary Hamiltonians for $\Psi$ and for the interaction between $\Psi$ and $\Theta$, as long as during the decay step $\Psi=0$ inside the device. This will be  true if we think of $\Psi$ as some sort of excitation (wave packet) having left the device by this time. Although this restricts the form of $H_{\rm{SL}}$, it has the advantage that it is no longer necessary to keep performing time-reversals once the training is finished and as long no evaluation passes are performed.  


Overall, what our training procedure (in either of the two cases) does is to impart an additional
effective force on $\Theta$ proportional to the gradient of the cost function, as we will explain in the next section. This breaks the degeneracy in such
a way that the only surviving stable fixed points are the minima of the cost function.

Up to now we have always considered the evaluation field $\Psi$ to adopt the shape of wave packets traveling through the device. However, even there, other possibilities exist: one could adopt a scenario where time-reversal of $\Psi$ is implemented entirely inside the device. In that case, we do not need to think of propagation along a particular direction, nor an architecture reminiscent of the structure of layered neural networks. Rather, the dynamics of $\Psi$ could resemble that of a wave-chaotic system inside a complex nonlinear cavity (see example in appendix D).

\subsection{Summarizing the general scheme of Hamiltonian Echo Backpropagation}
\label{subsec_summary}

\begin{figure}[h]
\includegraphics[width=8cm]{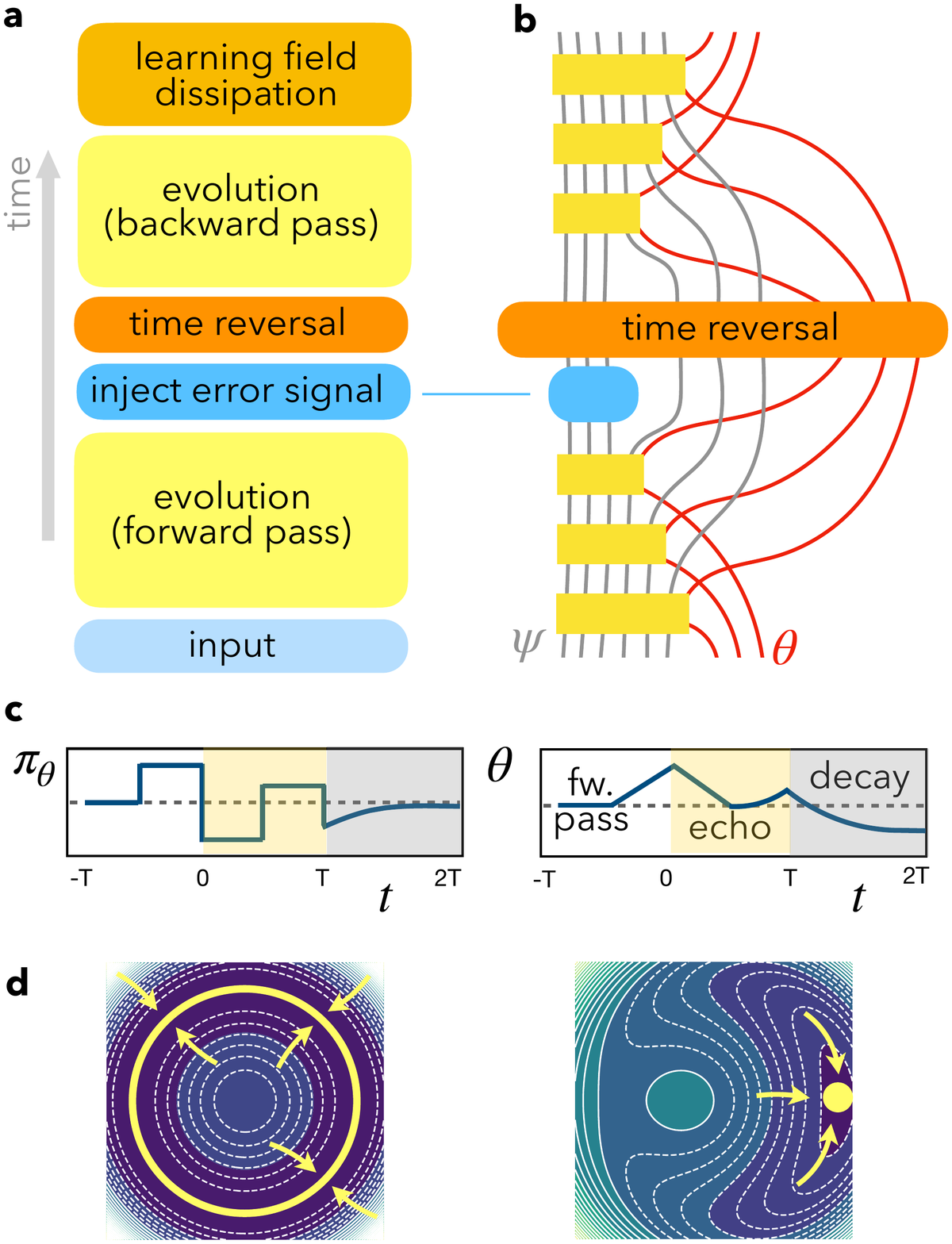}
\caption{\label{Fig4_Ingredients} \textbf{Ingredients of a self-learning machine based on Hamiltonian Echo Backpropagation.} (a) Sequence of events. (b) Use of 'pseudo-dissipation via ancillary modes'. The learning field $\Psi$ is here decomposed into a number of discrete modes. As time progresses, more and more modes do not participate any longer in the physical interaction (yellow boxes), becoming ancillary modes. Both these ancillary modes and the learning field $\theta$ also take part in time-reversal, however. (c) Time-evolution of the learning field momentum (left) and the field itself (right) during one forward-backward pass, including the effects of time-reversal (we show a single value of the field at one location). The momentum decays back to zero, but the field has been updated suitably. (d) The intrinsic dynamics of the learning field during the decay step has a continuously degenerate ground state (i.e. a manifold attractor). Right: The effect of the training dynamics is to break this degeneracy and produce a minimum determined by the cost function.}
\end{figure}

We are now in a position to summarize the general approach (Fig.~\ref{Fig4_Ingredients}). 
The HEB procedure consists of a sequence of multiple events (Fig.~\ref{Fig4_Ingredients}a), but they can be subdivided into two big steps:
\begin{enumerate}
\item \text{Echo step}. We
randomly draw an input sample  from the training data set, and we use it to inject $\Psi(-T)$. 
Following the nonlinear dynamics of the interacting system, the output is obtained at $t=0$. At that point, the evaluation field is weakly
perturbed according to $\Psi(0)\rightarrow\Psi(0)-i\epsilon\partial_{\Psi^{*}(0)} C$, which can be brought about via the dynamics induced by an extra interaction Hamiltonian (the "cost function Hamiltonian"), as explained above. Immediately afterwards, both $\Psi$ and also the learning field $\Theta$ are time-reversed (i.e. phase-conjugated). At $t=T$, a perturbed echo of the initial
configuration will be obtained. We then time-reverse $\Theta$ again at $t=T$.
The final result of this whole process is equivalent to the following
update of the learning field:
\begin{equation}
\Theta(T)=\Theta(-T)-i\epsilon\partial_{\Theta^*(-T)}C.\label{eq:update}
\end{equation}
To be clear, $C$ here always is the sample-specific cost function. We show in detail how to prove this result in general in the next
section. It has a simple interpretation: to first order in
$\epsilon$, the evolution of $\Theta$ satisfies Hamilton's equations,
with an effective Hamiltonian equal to the cost function.
\item \textbf{Decay step}. The dissipation is switched on for a time interval
$T_{D}$. The evolution
of $\Theta$ is determined by $H_{\rm{SL,decay}}$. This will ensure the suitable update of the $\Theta$ field needed to implement SGD via the physical procedure explained above. Finally, proceed with the next training sample.
\end{enumerate}

The evolution of the learning field during the whole procedure is schematically depicted in Fig.~\ref{Fig4_Ingredients}c. Imagine that these two steps are alternated many times, with suitable
small values for $\epsilon$ and $T_{D}$. Then the
average dynamics of $\Theta$ is approximately described by an effective Hamiltonian
given by

\begin{equation}
H_{\theta, {\rm eff}}=\frac{1}{\epsilon+T_{D}}\left(\epsilon C+T_{D}H'_{{\theta}}\right)\,,\label{eq:eff}
\end{equation}
where $C$ here represents the sample-averaged cost function, and $H'_{\theta}$ depends on the scenario we are considering. For case I (time-reversal of $\Theta$ outside the device), we simply have $H'_{\theta}=H_{\theta, \rm decay}$. For case II (time-reversal inside), we have $H'_{\theta}=H_{\theta}+H_{\theta, \rm decay}$. Remember that by our assumptions in both cases the phase
space associated to $H'_{\theta}$ has a continuous degenerate stable manifold. Consider what happens in
the low-temperature limit of the thermal dissipative reservoir that is used to act on $\Theta$. When we set $\epsilon>0$,
we break the degeneracy of $H'_{\theta}$ (see Fig.~\ref{Fig4_Ingredients}d). The dynamical
process we described will tend to equilibrate the temperatures of
$\Theta$ and the reservoir, so that $\Theta$ will tend to
minimize the cost function, constrained to the manifold of ground
states of $H'_{\theta}$. Once the training has converged,
we can stop perform training steps (in terms of the effective
Hamiltonian, that is equivalent to setting $\epsilon=0$). Since we
have converged to one of the many stable states, $\Theta$ will remain near the learned
configuration for arbitrarily long times. That is of course important
to ensure that, after training is completed, the SL machine doesn't
quickly forget the result of the learning process. This line of reasoning
does not depend on the particularities of $H'_{\theta}$,
as long as the phase space possesses some stable manifold.

\begin{figure}[h]
\includegraphics[width=8cm]{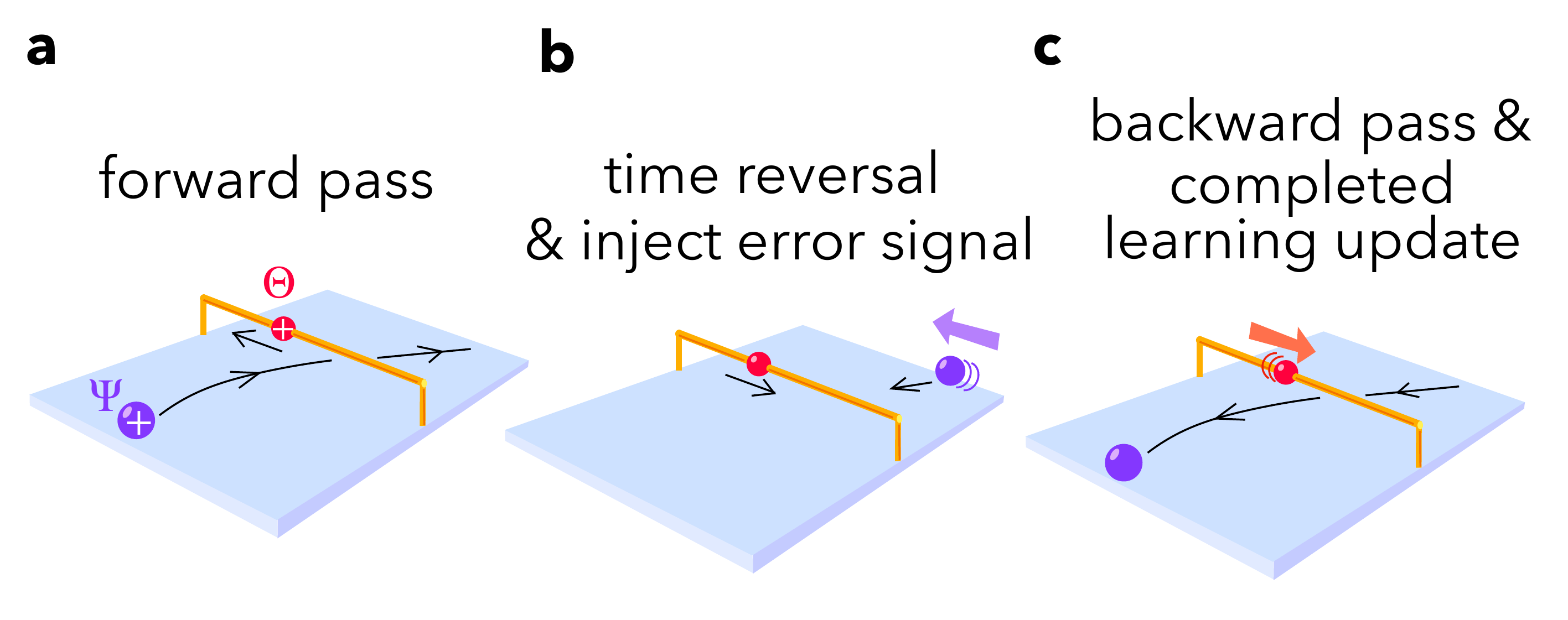}
\caption{\label{Fig_PinballSLM} \textbf{Hamiltonian Echo Backpropagation in a simple mechanical model.} 
(a) A charged ball $\Psi$ is launched with a velocity representing the input to the machine. Its trajectory is deflected depending on the position of another charged ball $\Theta$, which can freely move along a rail and which experiences a momentum kick due to this interaction. (b) Eventually, the velocities of both balls are reversed, and the ball $\Psi$ is slightly adjusted according to its deviation from the desired outcome ("error signal"). (c) Afterwards, it retraces almost exactly its initial trajectory, delivering yet another momentum kick to the $\Theta$ particle. Finally, any remaining velocity of $\Theta$ is dissipated and converted into a finite displacement, representing the training update. Of course, the expressive power of this machine, containing only these two degrees of freedom, is very limited (i.e. it can only produce very rough approximations to most input-output relations). Nevertheless, it illustrates the generality of the procedure -- e.g. an arbitrary (even unknown) static force field acting on the $\Psi$ particle will not spoil the learning procedure.
}
\end{figure}

This concludes our overview. The steps of the approach are illustrated in Fig.~\ref{Fig_PinballSLM} in terms of a very simple mechanical Hamiltonian system, highlighting the generality of the procedure.

We now present a few additional considerations.

Consider now what happens when the set of stable ground states of $H'_{\theta}$ is in fact not continuous, but actually discrete.
This may be the case if, for example, $\Theta$ describes a collection of
degenerate parametric oscillators. In this situation, a gradient descent strategy would not succeed, as $\Theta$ cannot change continuously between fixed points. Nevertheless, our procedure for training can still minimize the cost function. Indeed, consider that we start from a relatively high temperature in the thermal reservoir. The thermal fluctuations in the reservoir would result in   random jumps between the stable points. As our training procedure advances, we decrease
the temperature of the reservoir, inducing annealing. In the end, if
the system converges to a thermal equilibrium, the configuration of
$\Theta$ would be found in one of the minima of the cost function.

In practice, controllable dissipation may be a too stringent requirement. Actually, we do not need to have control over the dissipation on $\Theta$ if the dissipation timescale is large enough. In order to see this,
we require that the damping constant is weak enough that its effect
can be neglected during one single evaluation step: i.e. the damping
time-scale $\Gamma^{-1}$ is much longer than the duration of the
evaluation step $T$.  The core of our procedure is to enact an effective
force on $\theta$ that is proportional to the slope of the cost function.
If the dynamics were purely conservative, that would not be sufficient,
as $\Theta$ would oscillate around the stable manifold without ever
converging anywhere. It is clear then that we need such a damping term.
In consequence, we distinguish two time scales: the short time scale
$T$ of an evaluation step, in which all dynamics is approximately
Hamiltonian and time-reversible, and the long time scale of the total training time, in which dissipation ensures that, in the end, $\Theta$ converges
to the stable manifold.

\subsection{Invertible Dynamics vs. Contractive
Maps}


Before going any further, it is useful to introduce the concept of
\emph{pseudo-dissipation via ancillary modes}. The goal of supervised learning (both for
classification and regression) is often to approximate a map from
a set of high dimensional input vectors to a lower dimensional space.
This is an inherently contractive process. Nevertheless, the fact
that our theory of SL machines is based on reversible dynamics during the evaluation (forward pass) does
not imply that it cannot be used for this purpose. It is always possible
to embed contractive maps inside a higher-dimensional reversible map.
In physical terms, we can introduce pseudo-dissipation via ancillary modes: by introducing additional ancillary dynamical variables (which will also be considered part of $\Psi$ in our formalism), we can simulate the
effect of dissipation (Fig.~\ref{Fig4_Ingredients}b). We call it pseudo-dissipation via ancillary modes because, unlike
for real dissipation, we can still time-reverse the whole system,
including the ancillary modes.

Aside from its use to learn contractive maps, there is a further reason to include pseudo-dissipation via ancillary modes in SL machines. In order to be expressive, a SL machine must have a very complex nonlinear dynamics. However, it seems
advisable to avoid the strong sensitivity on the input that is implied
by chaotic dynamics.
When adding dissipation to a conservative chaotic system, it is known
that this reduces the sensitivity to initial conditions. In the case
of driven systems, where asymptotic Lyapunov exponents can
be defined, increasing dissipation leads to an overall shift of Lyapunov
exponents to smaller values, thus stabilizing the system \citep{hand_finch_1998}.
In the case of undriven systems, stronger dissipation enhances the rate for
leaving transient chaotic dynamics and settling into fixed points \citep{doublytransient}.
In our numerical experiments (reported later in \ref{Sec-NumericalExperiments}), we considered a photonic convolutional neural network with pseudo-dissipation. We find that,
consistent with these general observations, adding pseudo-dissipation helps to reduce
the sensitivity to initial conditions and improves the performance of the
self-learning machine.

Nevertheless, we shall remark that pseudo-contractive dynamics is not a requirement for HEB, as it can also train devices without such a property. Pseudo-contractive dynamics is only a desirable property in some interesting applications of a self-learning machine. However, there are other applications in which it is actually beneficial to have a reversible input-output relation for the relevant degrees of freedom, and hence no pseudo-contractive dynamics is required. For example, reversible residual networks have been recently found useful in image classification tasks \mbox{\citep{gomez2017reversible,chang2017reversible,behrmann2019invertible}}.

To avoid confusion, we emphasize that this conceptual split of $\Psi$ into the original evaluation field  and additional ancillary degrees of freedom is irrelevant with respect to the following analysis of the echo step. Therefore, to simplify the notation we just consider the full vector $\Psi$ without specifying how or whether it is split.

\section{Analysis of the Echo Step}

\label{Sec-Echo-Step-Math}

We now prove more formally that the crucial ingredient of our scheme,
the echo step, indeed works as advertised above, deriving equation
(\ref{eq:update}) in the most general setting.

 In order to simplify the notation, we define $\Phi:=\left(\begin{array}{c}
\Psi\\
\Theta
\end{array}\right)$
as the joint configuration of all dynamical variables of the system.
The dynamics of $\Phi$ is given by the following equation of motion

\begin{equation}
i\dot{\Phi}=\nabla_{\Phi^{*}}H,\label{eq:eqs_motion-1}
\end{equation}
where $H[\Phi]$ is the Hamiltonian. Crucially, we assume
$H$ to be time-reversal-invariant, which means $H[\Phi]=H[\Phi^{*}]$
in this notation. Let us remark here that time-reversal-invariance is the only assumption in this section (we do not need to assume a degenerate ground state for the expressions derived in this section). The operator $\nabla_{\Phi^{*}}$ involves the derivatives
$\partial/\partial\Phi_{i}^{*}$ and is formally defined as $(\nabla_{\Phi^{*}}f)_i = \partial f/\partial\Phi_{i}^{*}$.
We stress that we will, at the moment, study the dynamics
purely in the absence of coupling to the dissipative reservoir. In the
notation of Eq.~(\ref{eq:Hamiltonian_SL}), $H[\Phi]$
corresponds to the first three terms of the SL Hamiltonian, $H=H_{\psi}+H_{{\rm int}}+H_{\theta}$,
without the bath.

\subsection{Expression for the gradient of the cost function in terms of an 'advanced' backpropagating perturbation}

In a SL machine, the output of an evaluation step is a function of
the input $\Psi(-T)$ and the learning
field, whose configuration at the initial time is given by $\Theta(-T)$. In the
following, we want to obtain a formal expression for the gradient
of $C$ with respect to $\Theta(-T)$, which will later form
the basis for physical backpropagation. Since $\Psi$ and $\Theta$
are treated on an equal footing in Eq.~(\ref{eq:eqs_motion-1}), we
will set out to find the gradient of $C$ with respect to the initial
value of the joint state $\Phi(-T)$:

\begin{equation}
\begin{split}
\left(\begin{array}{c}
\nabla_{\Phi^{*}(-T)}C\\
\nabla_{\Phi(-T)}C
\end{array}\right)= D\left(\begin{array}{c}
\nabla_{\Phi^{*}(0)}C\\
\nabla_{\Phi(0)}C
\end{array}\right)\,,\label{eq:C_derivatives}
\end{split}
\end{equation} 
where the matrix $D$ is defined as 
\begin{equation}
D := 
\left(\begin{array}{cc}
\nabla_{\Phi^{*}(-T)}\otimes\Phi^{*}(0) & \nabla_{\Phi^{*}(-T)}\otimes\Phi(0)\\
\nabla_{\Phi(-T)}\otimes\Phi^{*}(0) & \nabla_{\Phi(-T)}\otimes\Phi(0)
\end{array}\right) 
\label{eq:D-definition}
\end{equation}and $\left[a\otimes b\right]_{ij}=a_{i}b_{j}$. Here the operator $D$ obviously maps back changes in the output to corresponding changes in the input, and we will now relate it to the Green's function of the problem. 

Suppose that $\Phi(t)$ is the solution of the nonlinear Hamiltonian
equations when the initial conditions are given by $\Phi(-T)$.
Let us consider what happens if we perturb $\Phi(-T)$ by some
small $\delta\Phi(-T)$. This results in a weak perturbation
$\delta\Phi(t)$ traveling forward on top of the zero-th order solution.
One can write $\delta\Phi(t)$ as the solution to the linearized equations
of motion:
\begin{equation}
L_{\Phi}(t)\begin{pmatrix}\delta\Phi(t)\\
\delta\Phi^{*}(t)
\end{pmatrix}=\delta(t+T)\begin{pmatrix}-i\delta\Phi(-T) \\
i\delta\Phi^{*} (-T)
\end{pmatrix}.\label{eq:linear}
\end{equation}
Here we introduced the linear differential operator 
$$L_{\Phi}(t)=
i\left(\begin{array}{c}
+1\\
-1
\end{array}\right)\frac{d}{dt}-\left(\begin{array}{cc}
\nabla_{\Phi}\otimes\nabla_{\Phi^{*}} & \nabla_{\Phi^{*}}\otimes\nabla_{\Phi^{*}}\\
\nabla_{\Phi}\otimes\nabla_{\Phi} & \nabla_{\Phi^{*}}\otimes\nabla_{\Phi}
\end{array}\right)H[\Phi(t)]$$
in compact notation, with $\left[\nabla_{\Phi}\otimes\nabla_{\Phi^{*}}\right]_{ij}=\frac{\partial^{2}}{\partial\Phi_{i}\partial\Phi_{j}^{*}}$
and so on. If the vector $\Phi$ has $M$ components, the operator
$L_{\Phi}(t)$ acts on a space of dimension $2M$. The inhomogeneous
term on the right-hand side of the preceding equation is there to enforce the initial conditions,
assuming that $\delta\Phi=0$ for $t\rightarrow-\infty$.

The solution of Eq.~(\ref{eq:linear}) can be written in terms of
the retarded Green's function associated to $L_{\Phi}[t]$:
\begin{equation}
\begin{pmatrix}\delta\Phi(t)\\
\delta\Phi^{*}(t)
\end{pmatrix}=\mathcal{G}_{\Phi}(t,-T)\begin{pmatrix}i\delta\Phi(-T)\\
-i\delta\Phi^{*}(-T)
\end{pmatrix}.
\label{eq:G_relation}
\end{equation}
The Green's function $\mathcal{G}_{\Phi}(t,-T)$ is a linear operator
such that 
\begin{equation}
L_{\Phi}(t')\mathcal{G}_{\Phi}(t',t)=\delta\left(t'-t\right)\mathbb{I},\label{eq:Greens_function_definition}
\end{equation}
where $\mathbb{I}$ is the identity matrix of size $2M$. Note that both $L_{\Phi}$ and 
$\mathcal{G}_{\Phi}(t,-T)$ depend on the zero-th order solution
$\Phi(t)$ of the nonlinear dynamical equations, around which we have
linearized. We chose $\mathcal{G}_{\Phi}(t,-T)$ to be the \emph{retarded}
Green's function, meaning that $\delta\Phi(t)$ originates at $t=-T$
and propagates in the forward direction of time. Causality dictates
that $\mathcal{G}_{\Phi}(t',t)=0$ for $t'<t$.

We now connect the matrix $D$ to the Green's function. From the definition of $D$ in Eq.~(\ref{eq:D-definition}), we see it measures the change at time $t=0$ when introducing a small change at time $t=-T$. The same relation is expressed by the Green's function, as defined in Eq.~ (\ref{eq:Greens_function_definition}), except for some constant factors. Merging both equations,
we see that

\begin{equation}
\mathcal{G}_{\Phi}(0,-T)=-D^{\dagger}i\sigma_{z}\,,\label{eq:Gret_vs_D}
\end{equation}
where $\sigma_{z}$ is the Pauli matrix. Therefore, we have $D=-i\sigma_{z}\mathcal{G}_{\Phi}(0,-T)^{\dagger}$,
and we can now express the change of the cost function in terms of
the Green's function. However, we will see presently that it is helpful
in terms of physical interpretation to introduce the \textit{advanced} Green's
function for this purpose, which also obeys Eq.~(\ref{eq:Greens_function_definition})
but has $\mathcal{G}^{{\rm adv}}(t',t)=0$ for $t'>t$ and represents
a signal going backward in time. It is related to the retarded Green's
function via

\begin{equation}
\mathcal{G}_{\Phi}^{\dagger}(t',t)=\mathcal{G}_{\Phi}^{\textrm{adv}}(t,t').\label{eq:Gret_vs_Gadv}
\end{equation}
Combining Eqs.~(\ref{eq:Gret_vs_Gadv}),~(\ref{eq:Gret_vs_D}) and
(\ref{eq:C_derivatives}), we finally obtain a formula relating the gradient of $C$ that is required for the learning update to the advanced Green's function and the (easily obtained) gradient of C with respect to the output fields:

\begin{equation}
\begin{pmatrix}\nabla_{\Phi^{*}(-T)}C\\
\nabla_{\Phi(-T)}C
\end{pmatrix}=-i\sigma_{z}\mathcal{G}_{\Phi}^{\textrm{adv}}(-T,0)\begin{pmatrix}\nabla_{\Phi^{*}(0)}C\\
\nabla_{\Phi(0)}C
\end{pmatrix}.\label{eq:gradient}
\end{equation}
This is still a formal expression at this point, but in the next section
we will show how it can be implemented physically.

\subsection{Time reversal of the perturbation: physical backpropagation}

How can we produce the signal of Eq.~(\ref{eq:gradient}) in practice, in order to allow the training update to take place? Apparently, we need to inject the source term $\nabla_{\Phi^{*}(0)}C$
as a perturbation and follow its "backwards time evolution".

Let us consider that, at $t=0$, we have obtained the output $\Phi_{{\rm out}}$.
Suppose that we have a way to implement an interaction Hamiltonian
$C[\Phi]$ that is proportional to the cost function (written
as a function of the output). Let us further suppose that we can control
this interaction Hamiltonian, so that we can switch it on during some
short period of time, and later switch it off again. In theory this
is always possible, because a cost function is just a real function
of the output. In practice, we will see that, depending on the cost
function, this corresponds to driving the system, or applying phase
shifts, or, for more complicated cost functions, to realizing some
nonlinear interactions. In any case, this operation is a function
only of the output, not requiring any knowledge of or feedback on
the internal degrees of freedom of the autonomous self-learning machine.
As we will discuss later, a practical implementation is possible for
several standard cost functions with relatively simple setups. For
the moment, we consider the general case in which $C$ is
an arbitrary cost function. When we realize such an interaction Hamiltonian
$C$, according to Hamilton's equations, the evolution will
be given by $i\dot{\Phi}=\nabla_{\Phi^{*}}C$. Let us suppose that we only realize
such interactions during an infinitesimally small time interval, $\epsilon$. 
The result of such a brief Hamiltonian evolution is to update the fields as
$\Phi(0)\mapsto\Phi(0)-i\epsilon\nabla_{\Phi^{*}}C$.

The advanced Green's function evolves a perturbation of the output
backwards in time to obtain the corresponding perturbation of the
input. In order to realize this physically, we need to induce the
time-reversed evolution of the system. In a time-reversal invariant
system (such as the one considered here), this is possible by implementing
a time-reversal operation and then letting the natural dynamics proceed
from there forwards in time. In the present complex notation, we know
that a time reversal is equivalent to phase conjugation, $\Phi\mapsto\Phi^{*}$,
flipping the momenta. By means of phase conjugation, applied to the
already perturbed field, we produce a perturbed echo given by $\Phi^{*}(0)+i\epsilon\nabla_{\Phi}C\equiv\Phi_{{\rm echo}}(0)+\delta\Phi_{{\rm echo}}(0)$.

Importantly, the whole configuration of the system is time-reversed,
and, as a consequence, the evolution for $t\in[0,T]$ is, to leading
order, given by the fully time-reversed nonlinear evolution: $\Phi_{{\rm echo}}(t)=\Phi^{*}(-t)$.
The first order correction $\delta\Phi_{\textrm{echo}}(t)$ is the
linear response of the system to the perturbation $i\epsilon\nabla_{\Phi}C$
in the initial conditions, evolving on top of $\Phi^{*}(-t)$.

Now we come to the crucial step. In any time-reversal invariant system, the advanced and retarded Green's functions of linear perturbations propagating on top of the original
nonlinear dynamics and its time-reversed counterpart, respectively,
are connected in the following way:

\begin{equation}
\mathcal{G}_{\Phi}^{{\rm adv}}(t',t)=\mathcal{T}\mathcal{G}_{\Phi_{{\rm echo}}}^{{\rm ret}}(-t',-t)\mathcal{T}\,,
\end{equation}
where $\mathcal{T}=\sigma_{x}$ is the time-reversal operation, interchanging
$\delta\Phi$ and $\delta\Phi^{*}$. Therefore:

\begin{equation}
\mathcal{G}_{\Phi}^{\textrm{adv}}(-T,0)\begin{pmatrix}\nabla_{\Phi^{*}(0)}C\\
\nabla_{\Phi(0)}C
\end{pmatrix}=\mathcal{T}\mathcal{G}_{\Phi_{{\rm echo}}}^{{\rm ret}}(T,0)\begin{pmatrix}\nabla_{\Phi(0)}C\\
\nabla_{\Phi^{*}(0)}C
\end{pmatrix}\label{eq:adv_backward_retarded_forward}
\end{equation}
It is now obvious that the left-hand-side, which (as we have already found above) formally yields the update needed for the gradient descent, can be \textit{physically} implemented in
terms of the right-hand-side version, with a suitable perturbation
injected and propagated \emph{forward} in time on top of the echo.
Finally, as we can see from Eq.~(\ref{eq:adv_backward_retarded_forward}),
at $t=T$ we have to perform a final time-reversal (phase-conjugation)
again.

Overall, after the whole process is completed, we obtain a replica
of the initial configuration at $t=-T$, but importantly with a small
correction that is proportional to $\epsilon$. The result of the
whole physical process can be summarized as 
\begin{equation}
\begin{pmatrix}\Phi(T)\\
\Phi^{*}(T)
\end{pmatrix}=\begin{pmatrix}\Phi(-T)\\
\Phi^{*}(-T)
\end{pmatrix}-\mathcal{T}\mathcal{G}_{\Phi_{{\rm echo}}}^{{\rm ret}}(T,0)\begin{pmatrix}\epsilon\nabla_{\Phi(0)}C\\
\epsilon\nabla_{\Phi^{*}(0)}C
\end{pmatrix}\,.
\end{equation}
We can convince ourselves that this is the learning update that was needed, by reformulating the deviation (second term on the right-hand-side)
with the help of Eq.~(\ref{eq:adv_backward_retarded_forward}) and
subsequently Eq.~(\ref{eq:gradient}){]}. This produces:

\begin{equation}
\Phi(T)-\Phi(-T)=-i\epsilon \nabla_{\Phi^{*}(-T)}C
\end{equation}
This is exactly the learning update that was needed, according to
Eq.~(\ref{eq:update}), for the learning parameters $\Theta$. Incidentally,
the evaluation degrees of freedom $\Psi$ have also changed, but they
will be discarded when the next input is injected into the learning
machine.

\section{Main Ingredients of  Implementations}

\label{Sec-IngredientsPhysicalImplementations}

   




The most important feature of our new method is its generality. It
can be used to train any time-reversible Hamiltonian physical system.
It is not our intention here to give a full description of all possible
implementations of the SL machine. On this general level, we cannot
even decide which ones are optimal, as the most suitable architecture is heavily dependent on the machine learning task of interest. Nevertheless,
we can give some guidelines.

The main ingredients of any SL machine of the type presented here are: a time-reversal-invariant Hamiltonian system with
two sets of dynamical variables $\Psi$ and $\Theta$; a device that can time-reverse $\Psi$ and $\Theta$, and a device or method
to transform the output according to the cost function Hamiltonian (injecting the error signal).  In this section, we will discuss each of these ingredients on a general level,
and we will provide a summary of some of the most promising possibilities for their realization.

\subsection{The nonlinear core}





As emphasized above, HEB can work with an arbitrary, completely unrestricted Hamiltonian, as long as time-reversal invariance is maintained. 

Any actual implementation will typically be based on local interactions between the degrees of freedom. Since for practical applications it is desirable to have designs that are scalable, a SL device in practice should be constructed by a combination of simple elements. We can think for example of integrated photonics or superconducting circuits. 

The details  of particular experimental platforms will be discussed later, in section \ref{sec:implementations}. Here, we briefly list some general guidelines.  In order to attain a nonlinear input-output relation that is expressive and useful for machine learning, we need that (1) many degrees of freedom interact during the forward pass and (2) there are nonlinear interactions whose effect is strong during such a pass. A common question in this regard would be the effect of chaos. Chaotic dynamics is defined as a property of the long-time limit, while here we are considering a finite evolution time for the forward pass. In practice, as we will see in the numerical examples, this is a matter of degree: it is useful to have sufficiently strong nonlinear effects, but not become too sensitive to small deviations in initial conditions.


A device combining linear local couplings and local nonlinearities is enough to produce extremely rich dynamics. For example, in optics one can use a combination of beam splitters and nonlinear layers to reproduce the architecture of a feedforward neural network \citep{shen_deep_2017}. However, there is no reason a priori to restrict to such a layout: Nonlinear wave fields interacting with each other inside some extended medium do not possess any obvious layered structure, but are still compatible with HEB, just like arbitrary lattices of coupled degrees of freedom.

One important requirement for HEB, is, of course, a time-reversal-invariant Hamiltonian. Practically, this means that dissipation in the course of the forward pass must be as low as possible. That already rules out some physical devices where a strong attenuation is unavoidable (e.g. when nonlinearities are based on absorption of waves). Nevertheless, even for promising platforms, such as coherent nonlinear optics, there is some remaining level of dissipation. In the limit of large scale networks, even a small amount of attenuation could be problematic, since its effect increases linearly with the propagation length. However, since we are considering classical devices, there is in principle no obstacle to employ amplification in order to compensate for the loss present in the device (we discuss the aspect of dissipation further when assessing concrete experimental platforms).





\subsection{The time-reversal operation}

\label{subsec-time-reversal-implementation}




The second key ingredient of HEB is the ability to perform a time-reversal operation after the forward pass. Such an operation does not represent purely Hamiltonian evolution, i.e. it does not preserve the Poisson brackets of coordinates and momenta. In practice, this means the operation necessarily requires dissipation.  As opposed to the case of quantum mechanics, in classical systems this requirement poses no fundamental obstacle. However, a physical and fast implementation is not necessarily straightforward in all classical systems. Fortunately, a well-known solution in the case of nonlinear optics \citep{boyd2012contemporary} is phase conjugation (for a review of this extended field, see \citep{he_optical_2002}). It  is by now well established that phase conjugation is not particular to optics. In fact, it can be engineered in any wave field where one can realize three- or four- wave mixing. For example, it has been demonstrated with acoustic waves \citep{brysev1998wave}, matter waves \citep{deng1999four}, microwaves \citep{chang1998microwave}, and spin waves \citep{serga2005parametric}. 

The fact that phase conjugation can be used to time-reverse wave pulses has long been understood (see e.g.  \citep{miller_time_1980} for the optical domain, and  \citep{martin_time_2008} for atomic matter waves).
The reversal of time evolution has been demonstrated in several nonlinear systems, including nonlinear optics \citep{noauthor_time-reversed_2013,sackey_kerr_2014}, electromagnetic wave-chaotic systems \citep{frazier_nonlinear_2013}, and water waves \citep{przadka_time_2012}.

Moreover, in systems where phase conjugation is not straightforward to achieve physically, it is often possible to measure the output and inject back a phase-conjugated replica. We note that this would replace one part of the HEB procedure by external processing and feedback, but the overall scheme could still be advantageous, since we still benefit from physical backpropagation and physical parameter update. This last possibility is know as digital phase conjugation, and it has been demonstrated with remarkable success in optical setups \citep{papadopoulos2012focusing}, using electro-optic components.

 
 



Although the concepts of implementations of phase conjugation are well-established, it may be helpful to recall the principle here. From a mathematical point of view, for any time-reversible Hamiltonian, the evolution can be reversed
by inverting the momenta: $\boldsymbol{\pi}\rightarrow-\boldsymbol{\pi}$.
In our complex notation, that means to perform a phase conjugation
of the fields: $\Phi\rightarrow\Phi^{*}$. As we remarked above, this step
cannot be realized with purely Hamiltonian evolution, not even by
breaking time-reversal symmetry momentarily. A time-reversal operation does not preserve the Poisson brackets, so it is not a canonical transformation. However, one can circumvent this problem by using ancillary modes.

For simplicity, we will now discuss a situation with discrete coupled modes of the wave field, the extension to extended waves is straightforward. Let us assume that we have a dynamical variable $\Phi$ that we
want to phase-conjugate. Additionally, we also have a second, ancillary
dynamical variable $\varXi$. An important assumption is that the
initial value of the ancilla is $\varXi_{\textrm{in}}=0$. The phase
conjugation can then be realized in three steps. First, we couple
$\Phi$ and $\Xi$ in such a way that, after a fixed time, we have
that $\Xi=\Phi_{\textrm{in}}^{*}$. This step can be realized with
Hamiltonian dynamics. For example, a Hamiltonian of the "parametric
interaction" form $H=\chi\left(\Phi\Xi+\Phi^{*}\Xi^{*}\right)$
would result in the desired outcome after an interaction time $t=\chi^{-1}\log\left(1+\sqrt{2}\right)$.
Second, we swap both modes, so that $\Phi=\Phi_{in}^{*}$. This step
can also be the result of Hamiltonian evolution, now choosing a "beam splitter Hamiltonian", $H=\chi\left(\Phi\Xi^{*}+\Phi^{*}\Xi\right)$.
Third, we erase the state of the ancilla (this is where dissipation enters), so that $\Xi=0$ again.



\subsection{Implementation of the decay step}
\label{sec:implementation-decay-step}






A third ingredient of our scheme is the ability to transform the physically backpropagated perturbation into an update of the learning degrees of freedom. More specifically, the goal of the decay step is to shift the learning degrees of freedom according to the momenta (which in turn are proportional to the gradient of the cost function), and at the same time dissipate (only) the momenta. Mathematically, we want to produce the following transformation
\begin{eqnarray}
    \label{decay_step_transform}
    \theta' = \theta - \eta \nabla_{\theta} C, \\
    \pi_{\theta}' = 0,
\end{eqnarray}
where $\theta'$ and $\pi_\theta'$ are the learning field quadratures after the decay step, and $\eta$ is the learning rate.

Importantly, this operation can be implemented by means of a sequence of three elementary transformations. Since all the most promising platforms we conceive are based on wavefields, we provide here an outline of its implementation for a wavefield mode representing a learning degree of freedom.  





The transformation given by equation (\ref{decay_step_transform}) can be approximated by a succession of three elementary steps. In the first step, we couple $\theta$ and $\pi_{\theta}$ by means of a phase shift. In the second step, $\Theta$ undergoes parametric amplification, a routine transformation in nonlinear optics and wave platforms in general. Finally, there is a step of attenuation. The overall transformation as a result of these three steps is given by
\begin{eqnarray}
    \label{decay_step_transform_1}
    \theta' = \sqrt{AG} \left(\cos(\varphi) \theta + \sin(\varphi) \pi_{\theta}\right), \\
    \pi'_{\theta} = \sqrt{\frac{A}{G}} \left(\cos(\varphi) \pi_{\theta} - \sin(\varphi) \theta\right),
\end{eqnarray}
where $\varphi$ is the phase shift angle, $G$ is the parametric gain and $A$ is the attenuation. 

 If we set $\sqrt{A/G} \ll 1$, we have $\pi_{\theta}\approx 0$. We need to compensate for the attenuation of $\theta$ by tuning the parametric amplification. By choosing $\cos(\varphi) \sqrt{AG}=1$, we attain the desired transformation given by equation (\ref{decay_step_transform}). Since $\pi_{\theta}=-\epsilon \nabla_\theta C$, the learning rate is then given by $\eta=\epsilon \tan(\varphi)$. Note that the learning rate can be controlled by tuning the phase shift $\varphi$ (and tuning the parametric gain accordingly). Importantly, the learning rate can be controlled independently of the size $\epsilon$ of the injected error signal, since we always have the free parameter $\varphi$ available. This is useful, since the values of $\epsilon$ should be neither too large (breakdown of the linearization approximation for the backpropagated error signal) nor too small (negative impact of extra noise).

This is not the only possible practical implementation of the decay step. Observe that the desired transformation can also be written in  complex form as $\Theta' = ( \exp^{-i\varphi}\Theta + \exp^{i\varphi}\Theta^*) / 2\cos(\varphi)$ (a straightforward calculation shows that the learning rate is again $\eta=\tan(\varphi)\epsilon$). It follows immediately that the decay step can be realized by a combination of  linear unitary operations (i.e. beam-splitters), phase conjugation and phase-insensitive amplification.  

Alternatively, observe that the transformation given by equation (\ref{decay_step_transform}) could be also obtained as the evolution of a mechanical degree of freedom $\theta$ with ballistic propagation under the influence of friction. In this case, the initial momentum $\pi_{\theta}$ is translated to a shift in $\theta$ (proportional to the inverse of the mass), while the momentum is dissipated in the long-term limit as a result of the friction. Therefore, the decay step is straightforward to implement in the case of mechanical degrees of freedom (the problem with mechanical implementations is that there is no straightforward method to realize the time-reversal operation, unless by measurement and feedback).

\subsection{The learning signal}






The cost function is the fourth essential ingredient in the design of a SL machine. In our prescription for HEB, one crucial step is to inject the learning signal by
perturbing the output fields according to $\Psi(0)-i\epsilon\nabla_{\Phi^{*}}C$,
where $C$ is a Hamiltonian
that is proportional to the cost function. For arbitrary cost functions,
this could always be achieved via some external processing and feedback, but as a general rule of course we want to avoid such steps.
For simple cost functions, it could be possible to engineer
a physical system whose evolution is actually described by the cost function Hamiltonian $C$.

In the particularly important case of the Mean Square Error (MSE) cost function, a physical
implementation is relatively easy. The MSE cost function is defined
as the squared difference between the actual output and the target output:
\begin{equation}
C[\Psi(0)]=\int dx\,g(x)\left|\Psi(x,0)-\Psi_{\textrm{target}}(x)\right|^{2}.
\end{equation}
Here $g(x)$ is a position-dependent weighting factor and will be zero in the areas that
correspond to the ancillary degrees of freedom (which will be time-reversed,
but not compared against any target). Expanding the square, we can
write the corresponding Hamiltonian  as $C[\Psi(0)]=g(x)\left|\Psi(0)\right|^{2}-2g(x){\rm Re}\left\{ \Psi(0)\Psi_{\textrm{target}}^{*}\right\} $
(where we have dropped $\left|\Psi_{\textrm{target}}\right|^{2}$,
which is just an irrelevant constant term). In terms of a physical
wave field, the first term, $g\left|\Psi(0)\right|^{2}$,
is just a shift in the frequency of the field. Its effect on $\Psi(0)$
is to induce a phase shift given by $e^{-i\epsilon g(x)}$. The only reason we do not completely neglect this phase shift is that $g(x)$ may be spatially inhomogeneous. The second
term, $-2gRe\left\{ \Psi(0)\Psi_{\textrm{target}}^{*}\right\} $,
is nothing but a drive proportional to the target. In other words, this term can be realized straightforwardly by injecting a weak copy of the target field distribution, adding it to the wave field $\Psi$, and realizing the corresponding phase shift set by $g(x)$.

Alternatively, the MSE cost function can also be realized using the
intensity of the field. Now, instead of forcing $\Psi(0)$
to match a target field, we only try to match a target intensity distribution.
In that case, the cost function Hamiltonian is given by $C[\Psi(0)]=g\left|\Psi(0)\right|^{4}-2g\left|\Psi(0)\right|^{2}I_{\textrm{target}}$.
In that case, the experimental realization of $C[\Psi(0)]$
is reduced to a homogeneous Kerr nonlinearity combined with a phase
mask proportional to the target intensity pattern.

Other cost functions may be realized with different interactions. Which one is optimal would depend on the problem at hand and how the output is encoded in the physical fields. In many cases, the choice of the cost function would not be a critical feature.

With regard to the amplitude $\epsilon$ of the injected perturbation ("error signal"), on the one hand it has to be small enough so that the weak perturbation is approximately in the linear regime. On the other hand, in a noisy device it is beneficial to have a larger amplitude, in order to overcome the effect of noise. Therefore, it is advisable to choose the amplitude of the perturbation as large as possible while remaining in the linearized regime. Once $\epsilon$ is chosen, the learning rate is not yet fixed: It can be controlled by changing the free parameters in the decay step. When the decay step and the error signal injection are performed outside the nonlinear core, there is in principle no restriction placed on those free parameters, and hence the learning rate can be tuned at will.


\begin{figure*}[ht]
\includegraphics[width=\textwidth]{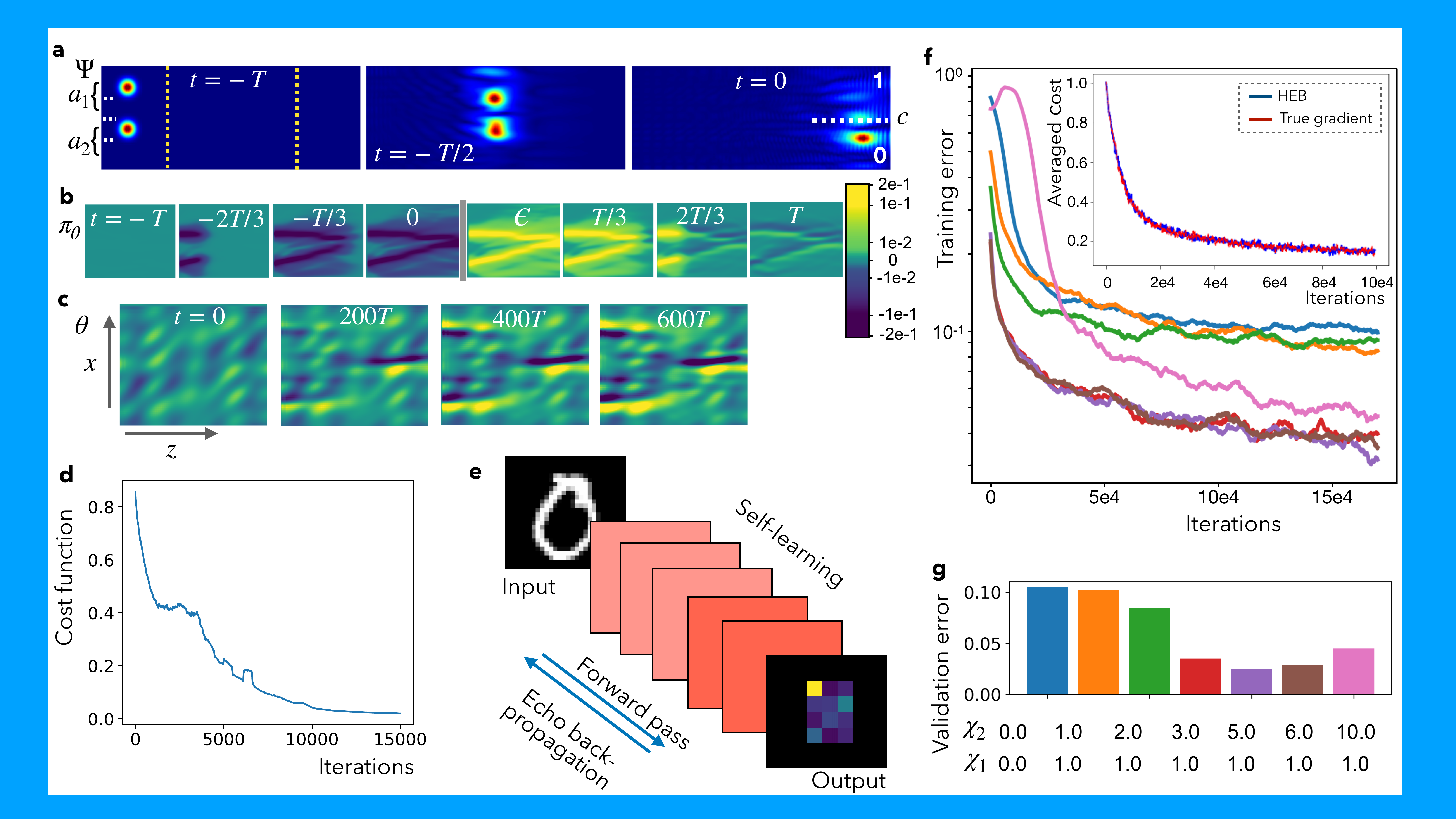}
\caption{\label{Fig5_Training} \textbf{Training a self-learning machine (simulations).} (a) Time-evolution of the "evaluation field" intensity $|\Psi|^2$ during a forward pass for a machine that has learned to implement XOR ($b=a_1\oplus a_2$).  The region in which the interaction with the learning field occurs is enclosed by yellow dashed lines. (b) Time evolution of the learning field momentum $\pi_{\theta}$ during a single  training step. The pattern builds up from left to right as the $\Psi$ pulse propagates. It is time reversed and then partially deleted as the $\Psi$ pulse travels back. The finite remainder at time $T$ is due to the injected error signal and produces the desired learning update. (c) Evolution of the learning field $\theta$  during many training steps. The training started with a smooth random configuration, and there are 600 HEB steps (each for a different randomly chosen training sample). (d) Cost function evolution during a particular training run for the model of (a)-(c).  (e) Setup of a self-learning machine that learns to classify handwritten digits (MNIST). The outcome is a one-hot encoding of the detected digit, i.e. one of 10 patches "lights up". (f) Training error evolution of the MNIST setup for different strengths of the wave field nonlinearity. The line colors represent a different value of the parameters $\chi_1$, $\chi_2$. $\chi_1$ and $\chi_2$ are the nonlinear strengths at the convolutional and dense layers, respectively. The color that represents each value of $\chi_1, \chi_2$  can be seen from (g). In the inset we display the cost function evolution when the SL machine is trained via HEB (blue line). We compare it with the gradient descent according to the true gradient (red line), with the same learning rate. The true gradient is obtained via autodifferentiation. In order to better compare the results, we average the cost function evolution over 10 different trajectories.} (g) Final validation set classification accuracy. 
\end{figure*}

\section{Illustration in numerical experiments}

\label{Sec-NumericalExperiments}

In this section we illustrate Hamiltonian Echo Backpropagation in
two specific examples, by means of numerical simulations (our simulations were performed using python and numpy). Both examples rely on nonlinear wave propagation, of the kind that might be implemented in photonic neural networks. In the first case, we consider a self-learning machine with a very simple setup and we train to learn a logical function.
In the second example, we consider a more elaborate architecture inspired by convolutional neural networks. We train the self-learning machine to perform classification on the standard MNIST handwritten digits data set. We believe
that a more sophisticated architecture or other choices of the activation
function and the cost function, together with a careful tuning of the hypeparameters, could improve the results even further
beyond the validation accuracy of 97.5\% observed here.
However, our primary goal here is to illustrate the concept, rather than constructing the optimal setup. In both scenarios, we simulated the full Hamiltonian Echo Backpropagation procedure, involving both the forward and the backward pass.

The setup considered in the first example consists of a nonlinear wave field. The dynamics of the evaluation field $\Psi$  is given by the nonlinear Schr\"odinger equation coupled to the learning field
\begin{equation}
i\dot{\Psi} = \frac{\beta}{2}\nabla^2\Psi + \left( \chi\theta+ g|\Psi|^2 \right) \Psi.
\end{equation}
The learning field $\Theta$ permeates the nonlinear medium and it interacts with the evaluation field through an interaction of the  $\chi^{(2)}$ type. This could be implemented in a nonlinear optical setup, or alternatively in a device based on matter waves. 
The evaluation field is initially prepared in a superposition of wave-packets that encode the input (Fig.~\ref{Fig5_Training}), with initial momentum $k_z > 0$. Thus, they propagate from left to right, accumulating the effect of the nonlinear self-interaction and the interaction with the learning field.  The output intensity is  measured at two points, $\mathbf{x}^{out}_{0},\mathbf{x}^{out}_{1}$. The output is considered a logical $0$ if $|\Psi(\mathbf{x}^{out}_0,0)|>|\Psi(\mathbf{x}^{out}_1,0)|$; otherwise, it is a logical $1$ (for more details, see Appendix \ref{Appendix-XOR}). The goal is to learn the exclusive-or (XOR) function mapping from input to output. As can be seen in Fig. \ref{Fig5_Training}, this is achieved rapidly.

The setup that we consider for the second example is inspired by the photonic neural network
introduced theoretically by Ong et al \citep{ong_photonic_2020}. We train this device to learn MNIST handwritten digits classification.
It is composed of three convolutional layers followed by a dense fully-connected
layer. The nonlinearity is provided by a $\chi^{(2)}$ optical nonlinearity,
as described in Eq. (\ref{eq:nonlinearitychi2}).
The convolutional
layers can be implemented by a combination
of  Discrete Fourier Transforms (DFTs) and learnable phase shifts. The proposal by Ong et al. did not address the problem of how to implement the dense layers; we introduce an implementation of the dense layers that is based on a sequence of DFTs, learnable phase shifts and a linear interaction with ancillary degrees of freedom (see  Appendix \ref{Appendix-CNN}). As a cost function, we choose the mean square error on the intensity.

In this setup, the learning parameters are all in the form of phase shifts. We assume that these phase shifts are produced by an interaction of the type $H_{int}=\theta\Psi^*\Psi$ inside some optical cavities through which the evaluation wave field travels.

In our simulations, the best results were obtained by choosing values of the nonlinear coupling strength that are different in the convolutional and the dense layers. This can be explained by the fact that the intensity decays as the wave packets propagate, as a result of the interaction with the ancilla modes. Therefore, it is optimal to choose that the nonlinearity in the dense layer, $\chi_2$, is stronger than the nonlinearity in the convolutional layers, $\chi_1$. The results in figure \ref{Fig5_Training} show the training progress for several values of $\chi_1$ and $\chi_2$. First we show the linear case, when $\chi_1=\chi_2=0$, and then we sweep across several values of $\chi_2$, with fixed $\chi_1$. One can see a sharp transition in the achievable accuracy when $\chi_2$ is increased. When the nonlinearity  surpasses some threshold ($\chi_2\approx 3$), the accuracy jumps from around 90\% to above 95\% (Fig.\ref{Fig5_Training}g). This result suggests that at this point the device gains a significantly better expressivity. Our observations are in qualitative agreement with the findings by Marcucci et al. \citep{marcucci_theory_2020}, where the expressive power of information processing using nonlinear wave fields was analyzed in terms of the strength of the nonlinearity. 

\begin{figure*}[t]
\includegraphics[width=\textwidth]{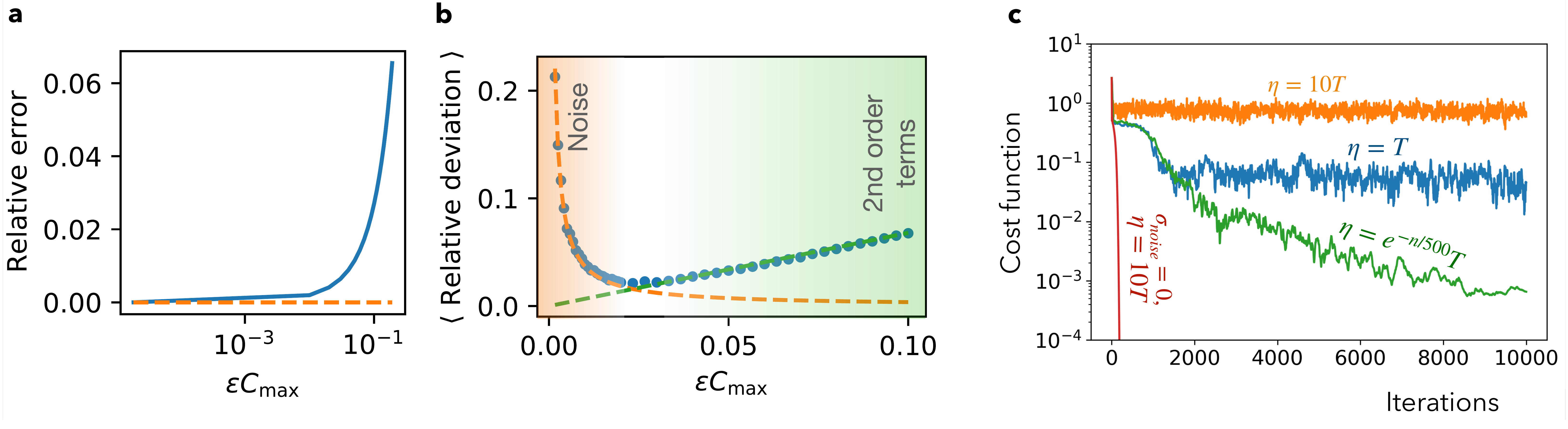}
\caption{\label{Fig_error_noise} \textbf{The effects of noise and nonlinear effects in the gradient estimation.} (a) The relative error in the approximation of $\partial_{\theta_j} C$, where $\theta_j$ is the value of the learning field in the middle of the evolution domain. We compare the update of $\theta_j$ obtained via HEB with an accurate estimation that was computed by means of the finite difference method. the result is presented as a function of the dimensionless quantity $\epsilon C_{\rm max}$, where $C_{\rm max}$ is the maximum possible value of the cost function.
(b) The root mean square error of the update obtained by means of HEB in the presence of noise. On the one hand, for small values of $\epsilon$, the noise overwhelms the term proportional to the gradient.  (c) Analyzing the influence of a noisy time-reversal operation on training. Here we consider a discretized version of the nonlinear Schr\"odinger equation, where the effect of noise is more prominent. We display the cost function during training for different values of the learning rate, at a fixed finite noise level whose standard deviation is given by $\sigma^2_{noise}=0.05I$ (see Appendix \ref{Appendix-XOR} for details).} 
\end{figure*}

We will now turn to study the effects of noise and nonlinear effects in the gradient estimation. We will do it for the case of the XOR scenario.

First, we studied the consequence of nonlinear effects in the gradient estimation - i.e. what happens for larger values of $\epsilon$. The  backpropagating signal is proportional to the gradient to first order in $\epsilon$, but for large values of $\epsilon$ the second order terms could be relevant. In figure \ref{Fig_error_noise}(a) we show the relative error in the update (with respect to the ideal gradient descent) as a function of $\epsilon$. The relative error is defined as $|\Delta \theta_j/\eta  - \partial_{\theta_j} C |/|\partial_{\theta_j} C|$, where $\Delta \theta_j$ is the update in one training step.

From the previous paragraph it would seem that it is optimal to use the smallest possible value of $\epsilon$. However, in a real self-learning device there may be noise (introduced during the time-reversal operation or other steps of the scheme). If $\epsilon$ is too small, noise will introduce significant errors into the gradient estimate provided by HEB. We analyze the situation in \ref{Fig_error_noise}(b). The root mean square error is defined as $\langle |\Delta \theta_j/\eta  - \partial_{\theta_j} C |^2/|\partial_{\theta_j} C|^2\rangle^{1/2}$, where the average is taken with respect to the noisy signal. On the other hand, for larger values of $\epsilon$, the previously analyzed nonlinear effects dominate. The orange and green dashed curves are a fit to the expected behaviour of the root mean square error in the limits of dominant noise ($1\epsilon$) and dominant nonlinear order effects ($\epsilon$).

We now analyze how to suppress the effects of noise. The operation of a SL machine is independent of the Hamiltonian, and therefore one does not even need to know the particularities of the device (including any possible deviations from intended fabrication) to make it work. This is true because the time-reversal operation produces an exact inversion of the Hamiltonian evolution, for any time-reversible Hamiltonian. One could then wonder how sensitive HEB becomes to the precision of the time reversal. In fact, since the nonlinear processes used to phase-conjugate signals can also amplify noise, there are practical limits to the performance of a time-reversal mirror. Nevertheless, we will show here that the effects of noise in the time-reversal can be mitigated, although at the price of a longer training time. In \ref{Fig_error_noise}(c), we compare the results using a  perfect time-reversal (red line) with the results obtained in the case of a noisy phase conjugation, for several values of the learning rate (orange and blue), as well as for a suitably chosen time-dependent learning rate ("learning rate schedule", an approach also known for ANNs, with $n$ denoting the training step index). As it could be expected, the performance of the SL machine deteriorates when  the time-reversal is noisy. However, one can observe that decreasing the learning rate improves the final accuracy. This is because the overall update after several steps will approximately cancel out the random noise, if the learning rate is small enough. Indeed, the backpropagating perturbation is a linear function of the error signal, to first order in $\epsilon$. Therefore, the update of $\Theta$ in each step will be the sum of the appropriate gradient update plus a random signal that is proportional to the noise in the time-reversal. If each update step does not change the value of $\Theta$ significantly, the overall change after several steps will average out the unwanted noise. Note that this behaviour only applies to temporally fluctuating noise. In case of static systematic deviations from the correct time-reversal operation, an independent calibration of that operation would be required. Importantly, however, that would not rely on any knowledge of the internal dynamics of the SL machine.

We note that in \ref{Fig_error_noise}(c) we studied a discretized version of the nonlinear wave field device considered above. The domain of the $\Psi$ and $\Theta$ fields is now a discrete lattice of size $20\times 30$ (we don't require that the fields are approximately continuous). The reason for this choice is that the cost function converges much closer to the optimal value. In this way, we can appreciate the effect of noise on convergence more clearly.

\section{Implementation Considerations and Potential Hardware Platforms}
\label{sec:implementations}


\label{Sec-PotentialHardwarePlatforms}

The elements for a SL machine that we have presented so far can be
realized in many platforms. In fact, many of the ingredients needed for
implementations of our proposal have already been demonstrated experimentally.
In some cases, the physical platforms that we consider below were
even already studied in the specific context of building new ML hardware, 
but it was not known how to train them sucessfully with physical procedures, without the use
of external computation and feedback. Hamiltonian Echo Backpropagation can now provide a means to do this. Our only aim here is to give a brief overview of the possibilities. It is clear that for each platform substantial further research will be required to come up with an optimized design respecting the particular hardware constraints, and to analyze the potential performance as well as to take steps towards first experimental realizations.

\subsection{General remarks on speed, energy efficiency, and scaling}

In general, neuromorphic devices are designed to offer intrinsic parallelism, which means the time needed for one evaluation becomes independent of the size of the input vector. HEB builds on the same advantage, with all the physical degrees of freedom (e.g. wave fields) evolving in parallel.

In many big-data applications, an important bottleneck for the speed of training may be the RAM bandwidth. For large ANNs, storing or retrieving the updated weights at each step can be the most important contribution to the training time. That points to a considerable potential advantage of training via HEB: the learning field is updated at the same time that the evaluation field propagates in the device, so there is no overhead associated to memory retrieval.

Whether a self-learning machine can be more energy efficient than existing systems would depend of course on the particular physical system used to implement it. It has been claimed by other authors that some physical neuromorphic platforms could be significantly more efficient than electronic devices. For example, in photonic integrated circuits the scaling of the power consumption can be more favorable \cite{wetzstein_inference_2020}. What we argue is that, if one uses one of those promising physical platforms, which compare favourably with digital devices, then training by means of HEB is a particularly efficient training method. For example, it is definitely more efficient than using a method based on parameter shifts and feedback. Indeed, applying the parameter shift method in a device with N parameters requires O(N) forward passes, while we only require O(1). 


There is a further reason why we believe that self-learning machines trained via HEB may be more efficient than electronic devices in the case of  large networks. In many of the most successful deep learning architectures, the same parameters are reused in many parts of the network. This can also be done in a physical learning machine of the type we envisage here, when the learning field interacts repeatedly with the evaluation field. This implies that the number of elementary operations scales faster than the number of parameters.  For example, a convolutional filter of size $M$ applied on an $N$ dimensional vector requires $O(MN)$ elementary operations. For $L$ layers, the number of elementary operations would thus scale as $O(LMN)$. While for an electronic device the energy consumption is roughly proportional to the number of elementary operations, the scaling of the energy consumption is different for a time-reversible machine. During the forward pass, the energy is conserved (the dissipation is ideally nonexistent or at least very low). The main energy consumption comes only from the time-reversal operation, the decay step, as well as the preparation of the initial states and injection of the error signal. The overall {\em unavoidable} energy consumption for a time-reversible machine thus scales like $O(N+M)$, where $N$ and $M$ are the dimension of the input vector and of the vector of learning parameters, respectively. Therefore, in general the scaling of the energy consumption as a function of the system size is more favorable in those cases where the number of elementary operations scales faster than the number of parameters.

\subsection{Nonlinear optical platforms}

Optical approaches to neuromorphic computing have recently attracted a lot of interest because they promise significant advantages in terms of speed, energy efficiency or parallelism \citep{guo_backpropagation_2020,hughes_training_2018,shen_deep_2017}. Importantly for us, those schemes involve a combination of coherent (linear and nonlinear) operations that are practically lossless, and hence they are amenable to our proposal. Optical SL machines would benefit from high parallelism and fast all-optical nonlinearities. 

In the nonlinear-optics approach to neuromorphic computing, on-chip photonic circuits are one of the most promising technologies for ML hardware. As discussed in section \ref{Sec-IngredientsPhysicalImplementations}, a scalable implementation of a nonlinear core would be combination of linear local couplings and local nonlinearities. A straightforward implementation of a self-learning machine would be to have both the evaluation field $\Psi$ and the learning field $\Theta$ be sent as pulses into an arrangement of waveguides, beamsplitters and cavities (some of them nonlinear), and to implement time-reversal outside this 'nonlinear core', as described in section \ref{Sec-IngredientsPhysicalImplementations}. The most direct implementation of a feedforward neural network would include layers of linear transformations alternated with nonlinear layers implementing the nonlinear function. In this line, there are several experimental demonstrations and theoretical proposals of deep neural networks based on programmable photonic circuits (e.g. \citep{shen_deep_2017}). The nonlinear activation functions could be provided, for example, by propagation inside a material with a strong Kerr effect, possibly enhanced via the use of resonators \citep{Yoshiki:14}.

The time-reversal operation could be implemented in any of the approaches discussed for fast phase conjugation in the nonlinear-optics literature; specifically by employing nonlinear optical interactions acting on the pulses propagating along waveguides or fibres, or, alternatively, inside nonlinear cavities with multiple interacting modes.

We have already discussed energy efficiency of HEB in general. Specifically for optical neural networks, it has been suggested \citep{shen_deep_2017} that they can be orders of magnitude more efficient in  terms of power than their electronic counterparts. An estimate of the power consumption would depend on the details of the particular implementation, but in any case the matrix multiplications and nonlinear activation functions are realized in the time-reversible, ideally lossless manner. The dissipative steps (phase conjugation, decay step, injection of input) require to dissipate (and/or resupply) an energy amount on the order of the energy contained in the optical pulses of the learning field and evaluation field. This follows from the way these fundamentally dissipative operations have to be implemented, as described before. We may estimate a typical energy on the order of 1 fJ  per wave packet, where we assumed the packet to be a 1 mW pulse of 1 ns duration, giving rise to sufficient nonlinearities in typical optical nonlinear resonators. The number of wave packets scales with the total number of learning parameters and size of the input vector, which gives an estimate for the total amount of energy dissipated per pass. 

With respect to the footprint of a photonic implementation, the limiting factor would be the size of elementary building blocks, such as beam splitters and resonators. If nonlinear effects are to take place during propagation along waveguides (instead of inside resonant cavities), then we also have to ask what is the minimum length to impart a strong nonlinear effect (e.g. a Kerr phase shift of $\pi$ on the wave fields), which is a function of the material, the waveguide design (mode localization), and the realistically feasible pulse energy. 

In nanophotonic circuits, first neuromorphic devices with an area of about $(100 
\mu m)^2$ per building block (tuneable beam splitter, in that case) have been demonstrated \citep{shen_deep_2017}, but future engineering should be able to scale down to around $(10 
\mu m)^2$, as suggested by previous developments (e.g. \citep{sun_large-scale_2013}). 3D integration may make it possible to realize large networks that are even more compact 
\citep{rechtsman_photonic_2013}.

We can also provide an estimate of the minimum feasible size of the nonlinear elements, based on the size of current all-optical switches. For example, a nonlinear microring resonator with a diameter of $70 \mu m$ was demonstrated to behave as an optical switch in the few mW level or less, with low dissipation \citep{Yoshiki:14}. With respect to the footprint of the components implementing decay step and time-reversal operation, we observe that their area scales only linearly with the number of degrees of freedom. Hence their footprint is not the dominant contribution in the large-scale (many-layer) limit. Additionally, as explained above, phase conjugating mirrors can be implemented by means of three- or four-wave mixing processes, for example inside a microring resonator \cite{sarapat2013conjugate}. Therefore, the size of the phase-conjugating mirror could be on the same order as a layer of nonlinearities. A similar argument applies to the decay step, which requires similar physical components.

With respect to the relevant time scales of a self-learning machine based on integrated photonics, we can also provide some rough estimates. The duration of a training step has four contributions: the forward pass, the time-reversal operation, the error injection, the backward pass, and the decay step. The time-scale of the forward pass (which has the same duration as the backward pass) will be proportional to the number of layers. Each layer would have a depth of $O(1)$ beamsplitters and nonlinear resonators. For beamsplitters with a size of $100\mu m$, the latency would be on the order of $10 ps$. In the case of the nonlinear microresonators, the relevant time-scale is roughly given by the inverse of their maximum operating bandwidth. For example, nonlinear microring resonators can operate with $100ps$ pulses \cite{turner2008ultra}. 

As we argued above for the footprint, the duration of the phase conjugation, decay step and error injection will not be the main contribution in the large-scale limit. Still, we can give some rough estimates. In the case of the time-reversal, as one example, optical phase conjugation has been demonstrated experimentally by means of a silicon  waveguide \cite{ayotte2007multichannel} having a latency of around $1ns$. However, it is reasonable to expect much faster phase conjugating elements, operating in the $100ps$ range, based on nonlinear microring resonators \citep{turner2008ultra}. With respect to the decay step, the limiting factor would probably be the step of parametric amplification. Since parametric amplification can be achieved by an adequate pumping of a three- or four- wave mixing process, we believe that the decay-step duration could be engineered to be on the same order as that of the phase conjugating mirror. Finally, the error injection corresponds to the injection of a signal and a phase shift element. Hence, the time-scale of the error injection will be comparable to that of a single linear layer, which was estimated to be on the order of $10ps$.

Like for any platform suitable for HEB, dissipation needs to be kept low, i.e. one needs to limit the attenuation of the optical fields. Substantial progress has been made in the area of integrated photonics in this regard, by careful materials selection, fabrication processes, and surface treatment as well as geometrical design. This has led to record achievements like the recently reported \citep{puckett2021422} ultralow-loss cavity in SiN with an optical quality factor of $4\cdot 10^6$, leading to a sub-MHz optical decay rate and relative losses less than $10^{-4}$ on the $100 {\rm ps}$ time scale. More generally speaking, we point out that Hamiltonian Echo Backpropagation in optical devices will profit from the current rapid push towards optical  quantum computing, since this involves the development of components with extremely low loss and strong nonlinearities, which are exactly the ingredients needed for our approach as well. Moreover, besides these ongoing improvements, attenuation could be compensated with coherent amplification (as we have mentioned before). In this way, one could place layers of amplifiers inside the integrated circuit, in such a way that the optical field is retained. Alternatively, the photonic device components could themselves be built as an amplifier (e.g. erbium-doped fibers) with a properly calibrated gain to compensate the loss. Similar to all the other aspects discussed above, further research will be needed, but the toolbox of nonlinear optics and integrated photonics is by now advanced and versatile enough for this challenge.

Another challenge in the case of optical implementations is the long-term storage of the learning degrees of freedom. In our approach, during each pass the learning field, besides undergoing time-reversals and the 'decay step', will have to be slightly amplified to counteract the small unavoidable losses. In this way, it is continuously kept alive throughout both the training and inference phases. However, eventually one will want to store it somehow, in order to keep persistence for longer times (indefinitely, instead of seconds or minutes). This could be done by performing a homodyne measurement on the learning field pulses and storing amplitude and phase on a digital computer. A more physical approach would involve holography. There are setups that have been proposed in the past \citep{wagner_multilayer_1987} that make use of holograms as a means to store the learning degrees of freedom of a photonic learning machine. In this way, very high storage densities could be attained \citep{psaltis_holography_1990}. Photorefractive holograms, e.g. based on lithium niobate, can be persistent for very extended periods, up to years.

Architectures beyond layered feedforward networks are also  realizable in integrated-photonics platforms. The basic ingredients of a recurrent neural network, i.e. a network with memory, can be implemented. One possibility would be to inject the inputs corresponding to a discrete time sequence as a train of wave packets, propagating along a waveguide. Assume that the waveguide is coupled to a ring resonator. If the pulse duration is much shorter than the ring's roundtrip travel time and the distance between pulses matches this time, then the subsequent pulses will overlap in the ring. When nonlinear interactions are present inside the ring, this is a way to obtain recurrent dynamics, where the output depends on the whole input sequence. Multiple such building blocks can be combined sequentially or in parallel, although we do not attempt here to work out a full working device. Training via HEB is possible even in this scenario, by time-reversing the whole train of pulses emerging out of the device.

Finally, besides integrated photonics, another interesting platform for optical HEB are possible. For example, the architecture proposed by Skinner et al. \citep{skinner_neural_1995} combines free propagation inside a crystal and a spatially homogeneous Kerr nonlinearity.  Another alternative may be so-called optical Ising machines. 
Ising machines have received a lot of attention in recent years
\citep{inagaki_coherent_2016,yamamoto_coherent_2017,pierangeli_large-scale_2019}. An Ising machine is an optical device that simulates the Ising Hamiltonian. One common implementation is based on a time-multiplexed train of pulses inside
a fiber loop with a degenerate optical parametric oscillator (OPO).
Due to the interaction with the OPO, the pulses have two degenerate
stable states that can be treated as spin-like variables. A linear
device in the fiber loop couples the pulses in a way that is reminiscent
of an Ising Hamiltonian. In this way, an artificial lattice can be
implemented with just a single fiber loop, e.g. demonstrating networks with 2000 degrees of freedom \citep{yamamoto_coherent_2017}. Although more research would be needed to work out a self-learning Ising machine, we can already now see that such a setup has all the elements
needed for our proposal: it has very complex nonlinear dynamics, it
is possible to engineer a degenerate ground state for $\Theta$, and
it can be time-reversed. As opposed to a standard photonic circuit, the number of nonlinear physical components can be strongly reduced, since they are re-used by the train of pulses.

\subsection{Superconducting  circuits}





Beyond optical circuits, superconducting circuits are another possible platform. In a somewhat similar fashion to photonic circuits, a typical superconducting circuit can be described as a 2D arrangement of coupled superconducting oscillators coupled by waveguides \citep{Devoret1997QuantumFI}. In fact, we can understand such superconducting oscillators as cavities that operate in the microwave regime. However, unlike photonic circuits, these superconducting cavities are made by coupling superconducting inductors, capacitors and Josephson junctions. 

In this scenario, our learning and evaluation fields are now microwave fields in the superconducting cavities. By using Josephson junctions, it is possible to engineer such superconducting cavities to induce strong non-dissipative nonlinear interactions. We should remark that here we consider that the circuit operates in the classical limit.

For the purpose of this paper, it suffices to understand that it is possible to  make lattices of coupled nonlinear superconducting cavities. In the end, this platform is not conceptually very different of the optical platforms described above, except for the fact that the $\Psi$ and $\Theta$ fields are now in the microwave spectrum, and the nonlinear elements are Josephson junctions.
In appendix \ref{sec-microwavecircuits_implementation_appendix}, we describe some of the design possibilities for constructing a self-learning machine based on this platform. Furthermore, we show some numerical results on the simulation of a self-learning device based on an array of nonlinear cavities, which could be implemented by means of a superconducting circuit. This example also illustrates the case of a self-learning device in which the learning field is always localized entirely inside the nonlinear core.

 The potential of superconducting circuits for fast and low-power classical computation has been recognized early on, which led to the development of so-called rapid single flux quantum logic in the 1990s.  Although the significant advances in more conventional electronics have resigned this to the status of a niche application, it has more recently been considered as fast control electronics for quantum processors. In addition, classical superconducting circuits are now being considered for realizing forms of artificial spiking neuron hardware \citep{schneider2022supermind}, e.g. based on superconducting nanowires \citep{toomey2020superconducting}, although these are designed to work in a dissipative regime. At the same time, the past 20 years have seen impressive advances in superconducting circuits operating in the quantum regime for quantum computation and simulation, building especially on superconducting microwave cavities that are made nonlinear by interaction with Josephson junction-based elements. Such applications have driven the significant minimization of losses, which is important for our purposes. Taken together, this means there now exists a very sophisticated toolbox for superconducting circuits that can be exploited for Hamiltonian Echo Backpropagation. 

Regarding their footprint, microwave cavities range from $cm$-scale (for 3D cavities and transmission line resonators) down to $(100 \mu m)^2$ in a lumped-element LC-circuit implementation (the Josephson elements are much smaller and can be neglected here). Even though overall we do not expect the same integration density as for optics, such circuits offer the attractive possibility to be integrated on-chip with quantum information processing devices, making use of the same technology. This would provide online self-learning on real devices with negligible latency and no signal distortion by propagation on cables to the outside. 

Coherence times in qubits now reach several $100 {\mu s}$, in cavities even up to $10 {\rm ms}$, while strong nonlinear effects at the {\em single-photon level} can be obtained during times of around $500 {\rm ns}$ or less. We intend operation in the classical regime, at much higher photon numbers $n$, where nonlinearities are enhanced, e.g. reducing the time needed for acquiring a Kerr-effect phase shift of $\pi$ by a factor $1/n$. Thus, in a conservative estimate, dissipation per nonlinear pi-phase-shift could be as low as $10^{-2}/n$, with $n$ easily reaching the order of $10^6$ or higher. This produces to an extremely good approximation the time-reversal-invariant Hamiltonian dynamics we are striving for, turning such systems into a very promising platform.




\subsection{Other approaches}

Approaches not based on optics or superconducting circuits are also possible, though we will only mention them briefly, and their detailed design would be a challenge for future research.

Clouds of cold atoms or molecules, possibly trapped in optical lattices, offer the possibility of storing information in the density distribution or the molecular orientation, making longer-term storage of the learning field $\Theta$ in principle less challenging than in the case of optical cavities. The interaction of an optical field $\Psi$ with the atomic or molecular cloud would also provide strong optical nonlinear self-interactions, as has been demonstrated with Rydberg atom clouds. Alternatively, the fields themselves could be realized as macroscopically populated, classical matter waves. This directly leads to the nonlinear Schr\"odinger equation, which we showed to be useful for HEB in our numerical examples. Four-wave mixing and phase conjugation with matter waves have been demonstrated early on \citep{deng1999four}. 

Likewise, spin waves share all the required ingredients for a SL machine.
We must remark that time-reversal symmetry must not be broken, which
means that spin-spin interactions may be acceptable, but not (for example)
light-magnon interactions via the Faraday effect. In the case of spin waves, there are already
experimental demonstrations of strong nonlinear interactions that
give rise to sigmoid-like activation functions \citep{wang_nonlinear_2020}, and phase conjugation by nonlinear interactions has been demonstrated \citep{serga2005parametric}.

\section{Conclusions}

In this work we have introduced a general method, Hamiltonian Echo Backpropagation, to train SL machines based on Hamiltonian physical systems. For any physical device that meets some general assumptions, our method is guaranteed to realize gradient descent optimization of a cost function. The HEB procedure updates the learning parameters autonomously and \textit{in situ}, leveraging all the advantages of a neuromorphic device. Whether the physical device could achieve a good performance on a particular task will depend on its size and its expressive power, but our method can always find a local minimum of the cost function via physical dynamics. Our numerical simulations illustrate how it can be applied both to simple physical systems and also to other more sophisticated architectures. In fact, we showed an example of its application in the case of a photonic convolutional neural network used for image recognition. Although there are many previous proposals and even experimental implementations of photonic learning machines, most of their potential advantages cannot be leveraged without a training method that is fully autonomous. We believe that when HEB is directly applied to such photonic devices, this will greatly enhance their capabilities.

Going beyond the realm of photonics, there are many other promising physical platforms for machine learning hardware, as we have mentioned. 
We believe that our method would be especially helpful when there is no control over the internal dynamics of the device. Indeed, since none of the steps in HEB depend on the underlying Hamiltonian, one can use the physical device as a black box. In this way, HEB is insensitive to device variability brought about by fabrication imperfections, and it can also cope well with noise (as we have shown in one of our numerical examples), which is particularly important in low-power applications with a suppressed signal-to-noise ratio. Our method not only allows to consider various physical platforms, it will also make easier to explore new architectures. As opposed to other previous proposals, HEB does not rely on a sequential multi-layered structure, which means that it can be used to train new architectures going beyond the structure of a standard ANN. 

The code used for the numerical examples is available on
Github \citep{gitlink}.

\subsection*{Acknowledgments}

We thank Dirk Englund and Stefan Krastanov for insightful discussions on some aspects of experimental implementations in the domain of integrated photonics.

\begin{appendices}

\section{Example: XOR function}
\label{Appendix-XOR}

The first example of a SL machine that we show has a particularly simple architecture. The evaluation dynamical variable is a nonlinear wave field propagating in a 2D square lattice. We consider a tight-binding model  with hard-wall boundary conditions in which every node of the lattice is coupled to its nearest neighbours. At each site of the lattice there is a nonlinear attractive Kerr self-interaction. The learning dynamical variable is given by another field that does not propagate. We assume that the learning field and the evaluation field interact through a term in the Hamiltonian given  by $\theta \Psi^*\Psi$, which is the simplest term that can produce local phase shifts.

During the time-reversible steps, the equations of motion for both fields are given by
\begin{equation}
i\dot{\Psi} = \frac{\beta}{2}\nabla^2\Psi + \left( \chi\theta+ g|\Psi|^2 \right) \Psi,
\end{equation}
\begin{equation}
i\dot{\Theta} = i\Omega\pi_\theta + \chi|\Psi|^2.
\end{equation}

The hard-wall boundary conditions are given by $\Psi(x=0,z)=\Psi(x=W,z)=\Psi(x,z=0)=\Psi(x,z=L)=0$. 

During the decay step, the evolution of $\theta$ is given by $\dot{\theta}= \Omega\pi_{\theta}$, $\dot{\pi}_{\theta}= -\Gamma\pi_{\theta}$.

The goal of this example is to learn a simple logical function. In particular, we choose the XOR function, $b=a_1\oplus a_2$, since it is well known that it cannot be learned by a single-layer perceptron. For each logical input, $(a_0,a_1)$, we provide a physical input as a superposition of wave packets, $\Psi(\mathbf{x},-T)= f^{0}(\mathbf{x},a_0) + f^{1}(\mathbf{x},a_1)$. Each wave packet, $f^s(\mathbf{x},a_s)$, is a gaussian wave packet centered around $((2s+1)W/8+a_s W/2,l_{in}),\,s\in\{0,1\}$ whose standard deviation is given by $\sigma$. In addition, each wave packet has an initial positive momentum in the propagation direction. In this way, each input boolean variable $a_s$  is encoded in the position of the center of the corresponding wave packet. 

We train such a self-learning machine to learn the XOR function using Hamiltonian echo back-propagation. 
The cost function is given by 
\begin{equation}
\begin{split}
C:=\int  dx\,dz\, \phi(\mathbf{x}) |\Psi(\mathbf{x},0)-\Psi_{tar}(\mathbf{x},b)|^2,
\end{split}
\end{equation} 
where $\phi(\mathbf{x},b)=\frac{1}{2\pi\sigma_{\phi}^2}. \sum_{j=0}^{1}  e^{-(\mathbf{x}-\mathbf{x}^{out}_{j})^2/(2\sigma_{\phi}^2)} $ is a sum of two Gaussian functions centered around $\mathbf{x}^{out}_0 = (w,0.8L)$ and $\mathbf{x}^{out}_1 =(3w,0.8L)$, respectively. The target field,  $\Psi_{tar}(\mathbf{x},b)$, is also a Gaussian function centered around $\mathbf{x}^{out}_b$. Therefore, our cost function is a version of the MSE weighted by $\phi(\mathbf{x})$. In this way, the SL machine tries to fit $\Psi(\mathbf{x},0)$ to $\Psi_{tar}(\mathbf{x})$inside the peaks of $phi(\mathbf{x})$, where the output is measured.
We finally evaluate the accuracy by the following prescription:  if $|\Psi(\mathbf{x}^{out}_{0},0)|>|\Psi(\mathbf{x}^{out}_{1},0)|$, the logical output is 0; otherwise it is 1.


In order to solve numerically the equations of motion for the time-reversible steps, we used a split-step method. In particular, we used the symmetrized Fourier split step method, which takes half a time step using the linear operator in momentum space, then takes a full-time step with the nonlinear operator in position space, and then takes a second half time step again with the linear operator in momentum space. In order to satisfy hard wall boundary conditions, the discrete sine transform was used. We approximate the effect of the cost function perturbation by a single small Euler step at $t=0$ (this is a reasonable approximation when $\epsilon$ is small enough).  For the decay step, we assumed that we are in the regime in which Eq.(\ref{eq_update_free}) applies. Therefore, for each HEB step we simulate a forward pass, a backward pass and then we update the parameters according to Eq.(\ref{eq_update_free}).

We used a square lattice with $50\times 150$ sites. The nonlinear strength was chosen as $g= 10^2 I^{-1}T^{-1}$, where $I:=(2\sigma)^{-2} \int d xdz|\Psi|^2 $ is a measure of the peak intensity of the input field. In this way, the accumulated phase shift due to the nonlinear self-interaction is much larger than $2\pi$. The dispersion coefficient has to be large enough to prevent a self-focusing collapse. This implies that $\frac{\beta}{2}\nabla^2\Psi$ should be at least on the same order as $g|\Psi|^2\Psi$. In our numerical simulation, we set $\beta=10^2\frac{(2\sigma)^2}{T}$.  With this choice of parameters, one can check in the numerical simulations that the extent of the wave packets remains in the order of $(2\sigma)^2$ during the whole forward pass. We choose the initial momentum of the wave packets as $k_z = \frac{3}{4}\frac{L}{T\beta}$, so that they propagate a distance $3L/4$ during the forward pass.

The parameter $\Omega$ determines the rate of change of $\theta$. In order to have results that can be more easily interpreted, we work in the regime in which $\theta$ remains almost constant during a single forward pass. Hence, we choose $\Omega=10^{-2}T$. Next, we must choose $\Gamma$ so that $\frac{\Omega}{\Gamma}\gg \frac{\partial C}{\partial \pi_\theta(-T)} \left( \frac{\partial C}{\partial \theta(-T)} \right)^{-1}$, if we want to work in the regime in which Eq.(\ref{eq_update_free}) applies.  Since the cost function depends on $\pi_\theta$ only indirectly, through the evolution of $\theta$, one can estimate that $\frac{\partial C}{\partial \pi_\theta(-T)}$ is on the order of $\Omega T \frac{\partial C}{\partial \theta(-T)}$. Hence, we must make sure that $1/\Gamma\gg T$. In particular, we set $\Gamma = 10^{-2}T$.  After one training step, the update in $\theta$ is of the order of $\epsilon\frac{\Omega}{\Gamma}\chi I$. Hence, the change in the phase shift in $\Psi$ due to the interaction with $\theta$ is in the order of $\epsilon\frac{\Omega}{\Gamma}\chi^2 IT$ . On the one hand, this update must be small enough so that the stochastic gradient descent works. On the other hand, it should not be so small that the training progress is too slow. We choose $\chi=10^2\left(\frac{\Omega}{\Gamma \epsilon I T}\right)^{1/2}$. In our numerical simulations, we set $I=1$, $T=1$, $L=3$, $\sigma=0.1$.

\section{Example: Image recognition}

\label{Appendix-CNN}

As an illustrative example of a SL machine, we considered a device based on a photonic neural network  proposed by  Ong et al \citep{ong_photonic_2020}, thereby showing how an existing device design can profit from our self-learning procedure. It is composed of three convolutional layers followed by a dense fully-connected
layer. The nonlinearity is provided by a $\chi^{(2)}$ optical nonlinearity,
as described in Eq. (\ref{eq:nonlinearitychi2}). Both the convolutional
layers and the fully connected layer can be implemented by a sequence
of discrete Fourier transforms (DFTs) and learnable phase shifts. In Ong et al., the activation function is a complex version of the ReLU function. Instead, we choose the activation function given by a $\chi^{(2)}$ nonlinear interaction, which in principle can be implemented in a physical setup. As a cost function, we choose the mean square error on the intensity.

The processing inside the photonic convolutional neural network that we consider is composed of a sequence of operations acting on a 2D square grid of $N\times N$ modes. In addition to the evaluation modes, there are learning and ancillary modes that take part in the sequential operations and also have to be time-reversed.

Any convolution (with periodic boundary conditions) is diagonalized by the DFT. Therefore, the weight matrix in a convolutional layer can be performed by applying a DFT, then a diagonal matrix, and finally the inverse DFT. In the case of a unitary convolution, the diagonal matrix is of the form $D_{jj}=e^{i\theta_j}$ (i.e., a phase mask). All these three operations are unitary operations that can be implemented with a lossless photonic circuit. The DFT matrix is of course fixed; all the learning parameters, $\theta_j$, are contained in the phase mask.

The last two layers of the network are fully connected, and they should be able to express any arbitrary complex matrix. The implementation of these fully connected layers was not descibed in \citep{ong_photonic_2020}, so we provide a possible version here. We use a decomposition based on the singular value decomposition. In the singular value decomposition, any complex matrix can be cast as a product of the form $USV$, where $U$, $V$ are unitary matrices and $S$ is a diagonal matrix.  The unitary matrices can always be expressed as a product of the form $U=D_0\prod FD_i $, where $F$ are DFT matrices and $D_i$ are diagonal unitary matrices (phase masks) \citep{pastor2021arbitrary}. Finally, the singular value matrix, $S$, can be implemented by means of a linear interaction with an ancilla, provided that the singular values are bounded by $S_{ii}\leq 1$. This is achieved by coupling every relevant mode to an ancilla mode by a means of a general $\textrm{SU}(2)$ linear transformation, when the ancilla is initially zero. Any such transformation can be achieved by means of the product of a 50-50 beam-splitter, a phase shift, and another 50-50 beam-splitter. Since an overall scaling factor in the weight matrix is unimportant, it is sufficient that our implementation can produce any complex matrix with $|S_{ii}|\leq 1$. 

All the trainable parameters are contained in the phase shifts, which in this example are realized by a local interaction Hamiltonian  given  by $\theta\Psi^*\Psi$. Just like in the first example, we assume that the machine operates in the limit in which the dissipation time-scale is much longer than the duration of a forward pass. 


For the numerical implementation, we solved the evolution of the evaluation and learning fields in a sequential manner. There are five kinds of operations inside each forward pass: (1) DFT's, (2) the interaction of $\Psi$ and $\Theta$, (3) the interaction of $\Psi$ and $\Psi_{anc}$ (the ancilla in the dense layers), (4) the nonlinear activation function, and (5) the cost function perturbation. Each of these steps represents a different layer in the photonic device, and therefore they can be solved separately. The 2D DFT was performed by means of a fast Fourier transform algorithm. The step (2) corresponds to the interaction described by the Hamiltonian $H=\chi_{l}\theta\Psi^*\Psi$. The equations of motion for this step can easily be solved analytically. Therefore, we can calculate the overall effect of the step (2) by means of a simple formula. In the same way, the step (3) is described by the Hamiltonian $H=J(\Psi^*\Psi_{anc}+\Psi\Psi^*_{anc})$, which can also be solved analytically. The step (4) is integrable too, with the solution given by  Eq.(\ref{eq:nonlinearitychi2}). The step (5) is a small perturbation that can be well approximated by a single Runge-Kutta step. For the backward pass, the same operations are performed, but starting with the phase conjugation of the output of the forward pass. Of course, one has to time-reverse not only the evaluation field, but also all the ancillary and learning variables.

The photonic convolutional neural network that we considered has three convolutional layers and two fully connected layers. The convolutional layers are somewhat unconventional: they only have one filter, but it has the same size of the output. As we already described, this is accomplished by an alternation of a DFT, a phase mask and another DFT. The first DFT acts on the full input image with $N^2$ modes, while the phase mask and the second DFT act only on the innermost $M^2$ modes. The rest of the modes are considered to be ancillary modes. In particular, the first layer has $(N,M)=(31,31)$, the second has $(N,M)=(31,21)$, and the third has $(N,M)=(21,17)$. The size of the input and output of the fully connected layers are $(17,9)$ and $(9,4)$, respectively. The input of the network is a rescaled version of the corresponding image from MNIST. The intensity of the field at each input mode goes from $0$ to $I_{max}$ (which has units of the square  root of an action). In the numerical simulation, we set $I_{max}=1$. Also, we set that all the elementary steps (activation functions, interaction with $\theta$, etc.) have duration $T_s=1$.  At the convolutional and dense layers we have two different values of the nonlinear strength, which are given  by $\chi_1$ and $\chi_2$ respectively. From Eq.(\ref{eq:nonlinearitychi2}) one can deduce that both $\chi_1$ and $\chi_2$  have to be chosen so that $\chi_{i}|\Psi|T_s\approx 1$ for the dynamics to be strongly nonlinear. The nonlinear strength was chosen as described in figure \ref{Fig5_Training}.  The coupling to the ancilla is given by $J=10^{1}T^{-1}_s$, since we want to ensure that the mixing of the ancilla and the relevant modes is on the order of 1. 

With regard to the decay step, it is assumed that we are in the regime in which Eq.(\ref{eq_update_free}) applies. Therefore, for each HEB step we simulate a forward pass, a backward pass and then we update the parameters according to Eq.(\ref{eq_update_free}). The coefficient of the cost function perturbation is taken to be $\epsilon=10^{-2}$, and we set $\Omega=10^{-1}T^{-1}$ so that the update of  $\theta$ during a single forward pass can be neglected. We use a damping coefficient given by $\Gamma = 10^{-1}T^{-1}$ so that the learning rate is given by $\eta=10^{-2}T$. Finally, by a similar argument to the learning of the XOR function, we set $\chi_l=10^1\left(\frac{\Omega}{\Gamma \epsilon I_{max} T}\right)^{1/2}$ (see  Appendix \ref{Appendix-XOR}).

\section{Implementation in superconducting microwave circuits}
\label{sec-microwavecircuits_implementation_appendix}

For our purposes, the simplest implementation would consist of a Josephson junction array, which is essentially a lattice of connected capacitors, inductors, and Josephson junctions. There is considerable design freedom, since one can choose the connectivity via the electronic wiring topology and the properties of the individual elements by parameters such as the junction area etc. Such an array implements a discrete nonlinear metamaterial. A classical Hamiltonian is obtained \footnote{M. Devoret, Quantum Fluctuations in Electrical Circuits, in Les Houches Lecture Notes of the 1995 school on Quantum Fluctuations, eds. S. Reynaud, E. Giacobino, and J. Zinn-Justin, Elsevier 1997} by introducing node fluxes $\phi_n$ via the time integrals of voltages across the circuit elements, and node charges $q_n$ via the time integrals of currents. These form canonically conjugate variables, with $q_n$ taking the role of momenta, such that the equations of motion for the circuit follow as $ \dot \phi_n = \partial H / \partial q_n$ and $ \dot q_n = - \partial H / \partial \phi_n$. The node fluxes are proportional to the superconducting condensate phase at each location in the circuit, and their difference across a Josephson junction indicates the phase difference that determines the energy. As an example, in a simple 1D array one can produce the Hamiltonian of a discrete sine-Gordon model, $H=\sum_n (2C)^{-1} q_n^2 - E_J \cos(\varphi_{n+1}-\varphi_{n})$, with $\varphi_n=(2 e / \hbar) \phi_n$. The capacitance $C$ and the Josephson energy $E_J$ characterize the circuit elements. Despite the appearance of Planck's quantum $\hbar$ in this formula, one can operate in the classical regime by ensuring $E_J \gg e^2/C$. More elaborate setups employ arrays of coupled microwave modes (produced using transmission line resonators, 3D cavities, or lumped-element LC circuits) which are modeled as LC oscillators in the Hamiltonian and which can be made nonlinear by the insertion of Josephson junctions. Expanding the $\cos$ term in the field amplitude yields a description of coupled modes analogous to nonlinear quantum optics (with Kerr terms etc.). Since the Hamiltonian is time-reversal invariant in the absence of external magnetic fields, this platform fulfills all the requirements for HEB, provided a suitable mechanism for the echo and decay steps is implemented, following our prescriptions in the main text \ref{Sec-IngredientsPhysicalImplementations}.

In a simple circuit of this type, the main conceptual difference in contrast to nonlinear optical platforms is the absence of a natural propagation direction, which makes the circuit a fully recurrent device in the language of neural networks (no layers, no feedforward processing). This could be circumvented by employing transmission lines as waveguides, coupled to nonlinear microwave cavities, resulting in a setup whose mathematical description is completely equivalent to that presented earlier for the nonlinear integrated photonics platform, only operating in a different frequency regime (and with different physical mechanisms for tunability, as well as considerably stronger nonlinearities). Nonlinear wave dynamics, and applications like amplifiers, have been experimentally demonstrated in that type of setup.

However, it would also be interesting to explore physical learning machines beyond the standard feedforward architecture. In fact, the idea of using nonlinear physical dynamics as an analog implementation of a recurrent neural network was already considered by T. W. Hughes et al. \citep{hughes_wave_2019}. Moreover, Tureci et al. \citep{tureci2021} already studied the use of a superconducting circuit as an analog implementation of a reservoir computer with recurrent architecture.  We take this opportunity to demonstrate the generality of our approach and consider deliberately a recurrent circuit. (see next appendix).

We argue that there are promising applications for a learning device such like the one we describe in the following appendix. For example, a similar learning device was proposed in a recent paper as a tool for efficient quantum state measurement \citep{tureci2021}. In their proposal, which is based on the ideas of reservoir computing, the physical degrees of freedom are not trained. Instead, they require external post-processing that has to be trained.  Our scheme could potentially allow to perform training \textit{in situ} in a more efficient way.

\section{Numerical example: Cavity array}
\label{sec:cavity-array-example}

In the main text, we have described two numerical examples, involving propagation of nonlinear wave fields that could be obtained either in a bulk medium or in integrated photonic circuits. Here, we will present numerical simulations for a third platform, namely cavity arrays. This is inspired by the superconducting microwave circuits (see previous appendix and discussion in main text), where it could be implemented.

In contrast to the examples considered in the main text, we note that this will require a mechanism to time-revert the fields inside all the cavities, and thus denies the treatement of the self-learning machine as an arbitrary 'black box', which had been possible if time-reversal and decay took place outside the device, as in the optical case. Nevertheless, given the great amount of engineering control available nowadays in these platforms, that is not a major constraint. Imagine a setup of coupled microwave cavities which have been made nonlinear by insertion of Josephson junctions and which can be driven parametrically by externally applied microwave drives (a standard scheme in manipulating such cavities, employed routinely for quantum information processing). This makes it possible to implement the crucial steps of our protocol: time-reversal, decay of the learning field, and injection of the error signal (as well as refreshing the input after one learning sample has been processed).

Consider the simplest kind of scenario, where an array of such nonlinear superconducting microwave cavities is arranged in a square lattice, with each cavity identical in the simplest case (though static disorder is not a problem, unless it affects the time-reversal or the decay step). In a coupled-mode description, we have the Hamiltonian

$$ H= - \sum_{m,n} J_{mn} a^*_m a_n + g \sum_n (a^*_n a_n)^2 $$

We distinguish between $\Psi$-cavities and $\Theta$-cavities, but their the only difference consists in the operations we apply to them: At the beginning, the field amplitude inside the $\Psi$-cavities is quenched to zero and then the training input is stored inside them. Afterwards, the evolution according to $H$ takes place. After a time-interval $T$, time reversal is applied to all cavities, using the kind of parametric setup discussed elsewhere in this manuscript. The error signal is injected, and after a further time $T$, the final decay step takes place.

\begin{figure*}[ht]
\includegraphics[width=\textwidth]{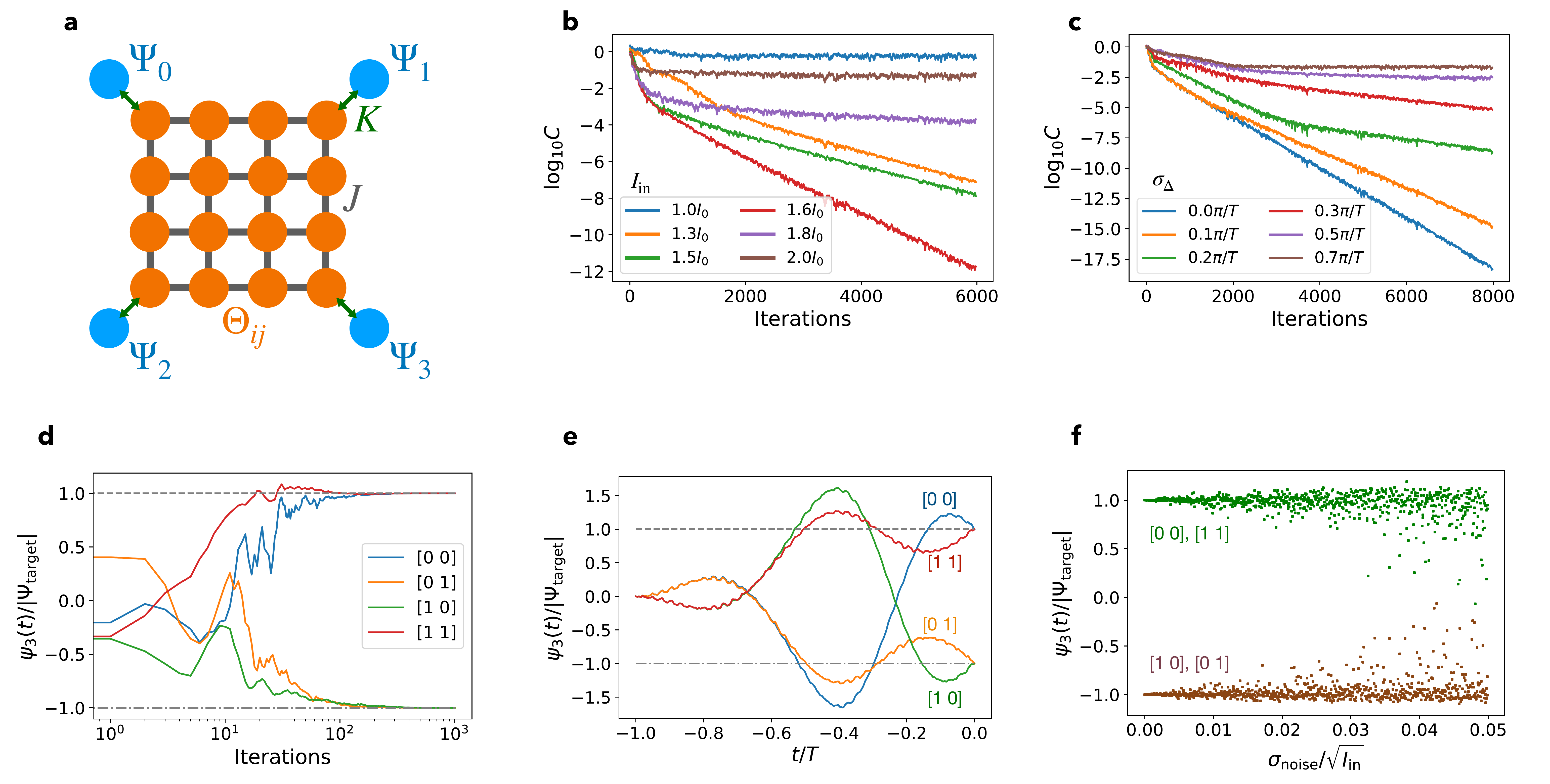}
\caption{\label{Fig_SupercondCavArray}  \textbf{Self-learning machine based on an array of nonlinear cavities (simulations)} (a) Schematics of the setup: the cavities that encode the learning degrees of freedom are placed on a 2D square lattice. Each nonlinear $\Theta_{ij}$ cavity is coupled to its nearest neighbours, with a coupling strength given by $J$. The cavities that represent the evaluation degrees of freedom, $\Psi_{i}$, are coupled to the $\Theta_{ij}$ cavities in the corners of the lattice, and the coupling strength is given by $K$. All cavities have an identical Kerr nonlinearity whose strength is given by $g$. The input is encoded in the initial values of $(\psi_{0}(-T),\psi_{1}(-T))$, while the output is given by $\psi_{3}(0)$. (b) Training error evolution for different values of the intensity of the input. We choose a reference level of intensity $I_0$ that fulfills  $gI_0 T = 3$. (c) In this plot we consider the robustness of the SL machine to fabrication imperfections. In particular we consider that each cavity has a random frequency shift whose distribution is given by $\mathcal{N}(0,\sigma_\Delta)$. Note that, since the random shifts are on the order of $0.1\pi/T-1 \pi/T$, their effect on the output value is very significant.  The HEB method can successfully implement gradient descent in the imperfect machine (since it is Hamiltonian independent). However, for larger values of $\sigma_\Delta$, the training process seems to converge to worse performance. Nevertheless, this result shows robustness to significant fabrication imperfections. (d) Evolution of the output as the training progresses (with an input intensity given by $I_{\rm in} = 1.6I_0$). Each curve represents each different input. (e) Time evolution of the field amplitude in the output cavity during a single forward pass, once the SL machine has been trained. (f) Outputs obtained when the input is noisy. We generated noisy inputs by adding Gaussian noise to the ideal input. After feeding each input to the SL machine, a noisy output is obtained. Here we plot the resulting outputs vs the standard deviation of the Gaussian noise. We can observe that the output remains highly concentrated around the target value when $\sigma_{\rm noise}$ is in the range of $0-3\%$ of the input amplitude. }  
\end{figure*}

We have numerically investigated this example, and again we trained it for the XOR training case, where only two input modes and two output modes are needed. The setup is explained in figure \ref{Fig_SupercondCavArray}(a). 

 The setup consists of $N^2$ cavities that encode $\Theta$ and 4 additional cavities that encode $\Psi$ (two input modes, two output modes).
The cavities corresponding to  $\Theta$ are placed in a $N\times N$ square array. The cavities corresponding to $\Psi$ are placed in the corners.
We set $N=4$. The Hamiltonian is given by a linear term that couples the neighbouring modes and a homogeneous local Kerr nonlinearity. To be precise, the linear term corresponds to a square tight-binding lattice.
    We work in a regime in which the ratio between the coupling strength of the tight binding Hamiltonian and $gI$ is very large: $J/(g I)\approx 10$, where $I$ is the average intensity of the $\Phi$ field. We empirically observe that in this regime the self-learning machine can train successfully. On the other hand, much larger nonlinearities often seem to lead to convergence issues. 
    
    We would not want to have that the output value is very sensitive to the specific timing of the readout of the output. For this reason, we also set that $K/J\approx 10$. In this way, the fast oscillations of the modes of the lattice are averaged out in the output.
    
    Due to the Kerr nonlinearity, the modes of the lattice will interact with each other via four-wave mixing processes. 
    We ensure that the interaction time is long enough that different modes of the lattice can completely mix with each other. In other words, we make sure that $gI T\gg 2\pi$.
    
    We simulate the dynamics by means of the Fourier split-step method. In figure \ref{Fig_SupercondCavArray}(b) we can observe that there is a finite range in the values of the input intensity for which the training is optimal (notice the logarithmic scale). In figure \ref{Fig_SupercondCavArray}(c) we study the effect of fabrication imperfections. In particular, we consider that the Hamiltonian of the device deviates from the ideal Hamiltonian by a random term that describes the fabrication imperfections. The random term in the Hamiltonian  corresponds to  random frequency shifts of each cavity mode in the lattice. The frequency shifts are drawn from a Gaussian distribution centered around $0$ and with deviation given by $\sigma_\Delta$. Crucially, HEB does not need knowledge of these random imperfections in order to realize backpropagation. We observe that the self-learning machine can minimize the cost function even for relatively large values of $\sigma_\Delta$, in the order of $0.3\pi T$. For very large values of $\sigma_\Delta$, even though it still performs gradient descent, the device seems to converge to poor local minima. Figures \ref{Fig_SupercondCavArray}(d-f) show the evolution of the output values during training, the evolution of the output during a single forward pass, and the effect of noise in the input.


\section{Ingredients of the Hamiltonian: Linear and nonlinear parts}

The dynamics of $\Psi$ and $\Theta$ is given by the Hamiltonian
$H_{\textrm{SL}}$. 
If one wants to use a SL machine for any useful task, the dynamics
of $\Psi$ must be complex and nonlinear. The expressive power of
ANN's can only be matched if the equations of motion for $\Psi$ incorporate
nonlinear interactions, as well as interactions with the learning
parameters encoded in $\Theta$. However, the input-output relation
must be deterministic and robust to small amounts of noise. These
considerations should guide the choice of the Hamiltonians. 

As we discuss later in this section, there is a considerable amount
of theoretical and experimental work about the time reversal of wave
fields, especially in nonlinear optics. For this reason, here and henceforth we will concentrate on
approaches based on nonlinear wave fields. Here, $\Psi(\mathbf{x})$
and $\Theta(\mathbf{x})$ are complex fields defined on a continuous
space, or in a discrete lattice. Although we make extensive use of
a notation that is reminiscent of optical fields, we must remember
that, actually, the setting is completely general (the real and imaginary parts
of the field are the two conjugate coordinates of a completely general
Hamiltonian degree of freedom). However, the fact that we have now assumed $\Psi(\mathbf{x})$ and $\Theta(\mathbf{x})$
to be fields allows us to make use of the notion of locality.

In a SL machine based on wave fields, $H_{\Psi}$ would
be composed of two parts: a \emph{free-field} Hamiltonian $H_{0}$,
and a second term $H_{\textrm{NL}}$ containing local self-interactions.
We can define a free-field\emph{ }Hamiltonian as any quadratic function
of $\Psi,\Psi^{*}$ and its time- and space-derivatives. More restrictively,
we can ask that $H_{0}$ is invariant with respect to a
global phase shift of $\Psi$, so that the total intensity, $\int d^{D}x\,|\Psi|^{2}$,
is a conserved charge. In that case, $H_{0}$ can only contain
second-order terms that are bilinear in $\Psi,\Psi^{*}$.

To produce non-linear dynamics, we introduce the self-interacting
term, $H_{\textrm{NL}}$. Often, this will be a local term.
Typical examples of self-interaction Hamiltonians would be the familiar
Kerr nonlinearity, $g|\Psi|^{4}$, or a Josephson nonlinearity, $\chi\cos\psi$, that can become relevant for superconducting circuit quantum electrodynamics.

The free-field Hamiltonian together with a self-interaction is enough
to obtain very complicated dynamics. But if we want to train the SL
machine, we still need to introduce some dependence on the learning
parameters through the interaction Hamiltonian, $H_{\textrm{int}}$.
In analogy with ANN's, we may ask that the interaction produces dynamics
linear in $\Psi$, although this is not a requirement. Moreover, $H_{\textrm{int}}$
will also be local in most realistic settings. Therefore, the most
reasonable choices for $H_{\textrm{int}}$ are $g\Theta^{*}\Theta\Psi^{*}\Psi$
(in nonlinear optics, this is called \emph{cross-phase modulation}),
or $g(\Theta+\Theta^{*})\Psi^{*}\Psi$ (the standard $\chi^{(2)}$ interaction in optics, or equivalently an optomechanical
interaction), or in general some interaction of the form $f(\Theta,\Theta^{*})\Psi^{*}\Psi$.
Alternatively, parametric interactions of the type $f(\Theta,\Theta^{*})\Psi^{*}\Psi^{*}+{\rm c.c.}$
could also be considered, leading to a different kind of dynamics,
although we will not discuss them here.

With respect to the strength of the nonlinear interactions, it should be large enough to attain a complex and highly expressive dynamics. However, some strongly nonlinear physical systems may exhibit chaotic behaviour. A system with strong sensitivity to the initial conditions is undesirable because any small perturbation can be strongly amplified, making impossible in practice to realize the time-reversed replica during the echo step. Ideally, in order to avoid strong sensitivity to the initial conditions while maximizing the expressivity of the device, the Lyapunov exponents should scale in such a way that the device is operated near the onset of chaos, but below it. Connected to this, a recent paper showed numerical evidence that some neuromorphic devices achieve optimal performance in the edge of chaos \citep{hochstetter_2021}. Other recent related works have studied how transient chaos can improve the expressive power of neural networks \citep{keup2021,poole2016,Farrell_2022}.

The Hamiltonian terms we described could be homogeneous across space,
like in the case of an optical field propagating inside a nonlinear
crystal, or they could be implemented using discrete building blocks,
as it would be the case for a photonic or microwave circuit with nonlinear
elements. It can be argued that the evolution of $\Psi$ under the
joint action of $H_{0}+H_{\textrm{int}}+H_{\textrm{NL}}$
is very similar to the evaluation of an ANN (as argued in other works before, see e.g. \cite{wagner_multilayer_1987,skinner_neural_1995,hughes_wave_2019}). On the one hand, the
effect of $H_{\textrm{NL}}$ is to induce a nonlinear activation
function. On the other hand, the part $H_{0}+H_{\textrm{int}}$
is quadratic in $\Psi$ and can be thought of as implementing the
trainable matrix of weights.

It is known that local controllable shifts together with fixed unitary
operations are enough to realize any unitary transformation \citep{reck_experimental_1994}. Of course,
it needs to be noted that standard ANNs can have arbitrary weight
matrices between layers, which seem to go beyond the unitary matrices
available in the present setting. However, any arbitrary matrix can
be embedded in a larger unitary matrix (with proper rescaling). For
this reason, the structure of the linear part of the ANN can always
be replicated in a SL machine. Alternatively, we may view an SL machine with unitary evolution as realizing a particular case of an invertible neural network.

\section{Pseudo-dissipation via ancillary modes}

As we said in an earlier section, pseudo-dissipation via ancillary modes can be a useful feature in a SL machine. We can imagine that in our SL machine
there are dynamical variables that are coupled to $\Psi$ until some
particular time, after which they evolve separately. Instead of explicitly
switching off the couplings as a function of time, we might also just
have wave fields propagating out of the region where interactions
are present. This can easily be arranged, e.g. in photonic circuits. The ancillary dynamical variables carry away part of the information, reconciling the contractive mapping with the overall time-reversal symmetry of the SL machine. We start with a high dimensional input, but as time passes, the dimension of the interacting part of the system is increasingly reduced. In the end, the dimension of the output may be much smaller than the original input dimension. Importantly, in the echo step of HEB, the output and the ancilla must be both time-reversed.

Pseudo-dissipation via ancillary modes via ancillary modes can  play an important
role for the activation functions implemented via nonlinearities.
As we already mentioned, the nonlinear evolution induced by the self-interaction Hamiltonian can be understood as an activation function. It is desirable to choose $H_{\textrm{NL}}$ in such a way that the induced
activation function is not oscillatory and has a bounded derivative.
Ideally, a monotonic function would be the best option, reproducing
standard ANN activation functions. In most previous approaches to
physical learning with nonlinear optical devices, the nonlinearity
is provided by a homogeneous Kerr self-interaction. The effect of
a Kerr nonlinearity is to induce a phase shift proportional to the
intensity of the field. Qualitatively, we can think of a Kerr medium
as a deep neural network where the activation function is of the form
$f(\Psi)=e^{ig\left|\Psi\right|^{2}}\Psi$. This is an oscillatory function, which is an undesirable property for an activation function in an ANN. What is worse, the derivative of the Kerr phase shift grows without bounds for increasing intensity. If we want to mimick the behaviour of a sigmoid or ReLU activation function, we must incorporate ancillary degrees of freedom, because all such standard activation functions are contractive. One possible choice is to use a $\chi^{(2)}$ nonlinearity, and to couple an ancillary mode to the evaluation field:
$H_{\textrm{SI}}=\chi^{(2)}\left(\Psi_{\textrm{anc}}^{*}\Psi^{2}+\Psi_{\textrm{anc}}\Psi^{*2}\right)$.
When the ancillary field is initially $\Psi_{\textrm{anc}}=0$, the
absolute value of $\Psi(t)$ as a function of the absolute value of
the input is qualitatively similar to a sigmoid function (although
it is not a monotonic function): 
\begin{equation}
\Psi(t)=\Psi(0)\textrm{sech}\left(2^{-1/2}\chi t|\Psi(0)|\right).\label{eq:nonlinearitychi2}
\end{equation}
Aside from $\chi^{(2)}$ nonlinearities, it is certainly also possible
to produce an adequate activation function by a judicious combination
of ancillary modes and a Kerr self-interaction Hamiltonian, or other
kinds of self-interaction. Sigmoid-like thresholders based on optical
Kerr effect have been previously described\citep{doran_nonlinear-optical_1988}, as well as bistable optical switches \citep{soljacic_optimal_2002}. It is known that the Kerr interaction in
combination with linear operations can be used to construct the Fredkin
gate, and therefore it is universal for classical reversible computation
\citep{kang_optical_2020}.

Nevertheless, we should emphasize that non-oscillatory activation functions are not a requirement for HEB, but a way to improve the expressivity of the device for those architectures that are more closely inspired on standard artificial neural networks. Just like some nonlinearities like ReLU work better in standard ANNs, in a SL device some nonlinear interactions have better results for some tasks. However, in general a SL device does not necessarily require non-oscillatory activation function, and in fact for many architectures of a SL device the concept of activation function does not apply (for example, consider a nonlinear core where the nonlinearity is homogeneously spread across all the device). Finding the best architecture for each task will involve some future experimentation, but in this section we offered in advance some insights about how to choose the best nonlinear systems.

\section{Stabilization of the learning field}

The primary method we envisage for HEB concerns wave fields being transmitted through a nonlinear core, and subsequent time-reversal operation and decay step outside this core. In that case, as explained in the main text, we are insensitive to the details of the Hamiltonian of the nonlinear core.

However, one can also envisage situations where the learning field is not constantly being time-reversed (during evaluation or training). When that is the case, one needs to make sure the learning field lives in a degenerate manifold, such that it is only the learning dynamics that will shift this learning field.

Once training via HEB has been completed and the cost function has
been minimized, the learning parameters should remain stable for long
times. The problem then is how to choose a physical system with degrees of freedom that can remain stable in a continuum of states. Mechanical degrees of freedom would be a priori a good selection. The position of freely moving nanoparticles or the orientation of a (possibly levitated) rotor (e.g. a molecule or nanorod) are examples
of systems that could store a degree of freedom for indefinitely long
times. The problem is that the time-reversal of mechanical systems
has no straightforward solution. The possible exception to this statement are optomechanical systems, where the coupling to the light field can be employed to time-reverse mechanical motion, but then the mechanical degree of freedom oscillates around a single equilibrium position, making it unsuitable for our purposes absent any additional ingredients. By contrast, in the case of wave fields, the experimental techniques for phase conjugation are better known, but storing information for longer times is more problematic.

The solution is to engineer a suitable physical system that has the desired characteristics. Modern systems in integrated nonlinear optics and in superconducting microwave circuits allow for the design of a wide range of nonlinear interactions between localized modes of the radiation field. In the following, we provide two examples of how this flexibility might be exploited to design a degenerate ground state manifold for a learning field.

In the first example, we consider a cavity that has two
modes $\alpha_{1,}\alpha_{2}$ with the same frequency (like counterpropagating
modes in a microresonator). Inside the cavity there is a nonlinear
material that induces a repulsive Kerr-type interaction. Both modes
have the same lifetime. In order to have stable states with non-zero
amplitude, we couple both modes to a parametric drive. Let us
define the quadratures of the field modes as $q_{1,2}={\rm Re}\{\alpha_{1,2}\}/\sqrt{2}$,
$p_{1,2}={\rm Im}\{\alpha_{1,2}\}/\sqrt{2}$ and introduce the abbreviations
$Q^{2}:=q_{1}^{2}+q_{2}^{2}$, $P^{2}:=p_{1}^{2}+p_{2}^{2}$. In terms of the quadratures, the full Hamiltonian reads 
\begin{equation}
H=-\chi\beta Q^{2}+gQ^{4}+\left(\chi\beta+2gQ^{2}\right)P^{2}+gP^{4}.
\end{equation}
This Hamiltonian is written in the frame rotating at the mode frequency,
and we assumed the parametric drive amplitude $\beta$ is purely real-valued.
It is not hard to see that this Hamiltonian is qualitatively similar
to that of a point particle moving inside a 'Mexican hat' potential.
Using this analogy, it is clear that any state with $P=0$ and $Q^{2}=\chi\beta/(2g)$ is a ground state. We could consider the angle in the $(q_{1},q_{2})$-plane as our learning parameter $\theta$. In order to couple to this parameter, the simplest option we found is to suppose that $\Psi$ interacts
differently with $\alpha_{1}$ and $\alpha_{2}$. For example, let us consider that there is a nonlinear element that couples $\Psi$ only to $\alpha_{1}$, through a cross-phase modulation. The interaction Hamiltonian then can be written in terms of the learning parameter as $H_{\textrm{int}}=g\cos^{2}\theta\Psi^{*}\Psi$.

Of course, storing a learning parameter in the amplitudes of parametrically driven
Kerr oscillators is not the only possibility. As a further example,
we briefly present another setup in which $\theta$ is the relative phase
between two modes. Consider two modes $\alpha_{1}$, $\alpha_{2}$
that are also parametrically driven. The two modes $\alpha_{1}$,
$\alpha_{2}$ have different frequencies given by $\omega_{1}$, $\omega_{2}$,
respectively. This physical system can be modelled by a Hamiltonian
given by $\chi^{(2)}\left(\beta\alpha_{1}^{*}\alpha_{2}+\beta^{*}\alpha_{1}\alpha_{2}^{*}\right)+\sum_{j=1}^{2}\omega_{j}\alpha_{j}^{*}\alpha_{j}$.
It is clear that this Hamiltonian is invariant with respect to any
global phase shift $\alpha_{j}\rightarrow e^{i\varphi}\alpha_{j},\,j=1,2$.
Let us introduce the notation $\alpha_{j}\equiv\left|\alpha_{j}\right|e^{i\varphi_{j}}$,
where $\varphi_{1},\varphi_{2}$ are the phases of $\alpha_{1}$,
$\alpha_{2}$, respectively. We define the learning parameter as $\theta:=\varphi_{1}+\varphi_{2}$.
Because of the symmetry of the Hamiltonian, $\theta$ is a free parameter
in the ground state. The challenge is now to find a way to couple
$\Psi$ to the learning parameters. One possible way is to consider
that $\Psi$ has two components (field modes) $\Psi_{1}$, $\Psi_{2}$.
Choosing adequately the frequency of the modes $\Psi_{1}$, $\Psi_{2}$,
the product $g\alpha_{1}\alpha_{2}\Psi_{1}^{*}\Psi_{2}+c.c.$ can
be made resonant. Then, $\theta$ parametrizes a linear interaction
between the two components of $\Psi$. This requires frequency stabilization
of the difference frequency of $\Psi_{1}$ and $\Psi_{2}$ vs. the
sum frequency of $\alpha_{1}$ and $\alpha_{2}$. The advantage, though,
is that now the symmetry of the Hamiltonian is more robust.

As we illustrated above, in order to engineer a physical system that
can store a continuous learning parameter, we need to realize a Hamiltonian
with some continuous symmetry. The problem is that in any real implementation
we may expect that any continuous symmetry will be broken at some
energy scale. If that is the case, most configurations of the learning
field after training will have a finite lifetime. Remember that we
assume that, after training, the learning field ends in one of the
degenerate ground states of $H'_{\theta}$, which is an ideal, exactly
degenerate Hamiltonian. If the true Hamiltonian of the system is instead
given by $H'_{\theta}+\delta H'$, the learning field
will slide down in the energy landscape of $\delta H'$. Eventually,
it will end up very far from the configuration that had been obtained
as a result of training, unless the training is periodically repeated to stabilize the "memory" of the system.

That potential problem could be avoided if the learning parameters
were stored in a discrete set of states. Consider that now $H'_{\theta}$
possesses a discrete set of degenerate ground states. In such a case,
a small perturbation in $H'_{\theta}$ does not affect the
long term stability of $\theta$, as long as it is much smaller than
the energy barrier between the degenerate states. Stochastic gradient
descent (SGD) is a strategy used to optimize continuous degrees of
freedom. Although HEB is originally motivated by the search of a physical
realization of SGD, we already argued that it can still be useful
for training discrete parameters. In that case, HEB works in an analogous
fashion to thermal annealing. In this picture, one can think of $\theta$
as a spin system that can be approximately described by a thermal
distribution. The probability of each state is then given by $e^{-\beta H_{\textrm{eff}}[\Theta]}$,
where $H_{\textrm{eff}}[\Theta]$ is the effective Hamiltonian
of Eq. (\ref{eq:eff}). If the temperature is slowly reduced, the
system will undergo annealing, and $\Theta$ would end in one of the
local minima of $H_{\textrm{eff}}[\Theta]$. Spin-like degrees
of freedom can be realized, for example, with Optical Parametric Oscillators \citep{inagaki_coherent_2016}.

\end{appendices}

\bibliography{SelfLearning}

\end{document}